\documentclass[10pt,twocolumn,letterpaper]{article}

\usepackage{arxiv}

\usepackage[utf8]{inputenc} 
\usepackage[T1]{fontenc}    
\usepackage{url}            
\usepackage{booktabs}       
\usepackage{amsfonts}       
\usepackage{nicefrac}       
\usepackage{microtype}      
\usepackage{xcolor}         

\usepackage{graphicx}
\usepackage{booktabs}
\usepackage[utf8]{inputenc} 
\usepackage[T1]{fontenc}    
\usepackage{url}            
\usepackage{nicefrac}       
\usepackage{microtype}      
\usepackage{bbding}
\usepackage{pifont}
\usepackage{bbm}
\usepackage{bm} 
\usepackage{soul}
\usepackage{multirow}
\usepackage{multicol}
\usepackage{makecell}
\usepackage{framed}

\usepackage{amsmath}
\usepackage{amssymb}
\usepackage{mathtools}
\usepackage{amsthm}
\allowdisplaybreaks[4]

\usepackage{wrapfig}
\usepackage{caption}

\usepackage{amsmath}
\usepackage{amssymb}
\usepackage{mathtools}
\usepackage{amsthm}

\theoremstyle{plain}
\newtheorem{theorem}{Theorem}[section]

\theoremstyle{definition}
\newtheorem{definition}[theorem]{Definition}

\theoremstyle{remark}

\usepackage[textsize=tiny]{todonotes}

\usepackage{natbib}

\usepackage[pagebackref,breaklinks,colorlinks]{hyperref}

\usepackage{amsmath}
\usepackage{amssymb}
\usepackage{mathtools}
\usepackage{amsthm}
\usepackage{bbold}
\newtheorem{hypothesis}{Hypothesis}
\usepackage[capitalize,noabbrev]{cleveref}

\begin{document}
\theoremstyle{plain}
\theoremstyle{definition}
\theoremstyle{remark}
\newtheorem{phenomenon}{Phenomenon}

\title{Randomness of Low-Layer Parameters Determines Confusing Samples in Terms of Interaction Representations of a DNN}

\author{
    \textbf{Junpeng Zhang}\textsuperscript{1, 2} \quad
    \textbf{Lei Cheng}\textsuperscript{1} \quad
    \textbf{Qing Li}\textsuperscript{2} \quad
    \textbf{Liang Lin}\textsuperscript{3} \quad
    \textbf{Quanshi Zhang}\textsuperscript{1}\thanks{Quanshi Zhang is the corresponding author. He is with the Department of Computer Science and Engineering, the John Hopcroft Center, at the Shanghai Jiao Tong University, China.} \\
    \textsuperscript{1}Shanghai Jiao Tong University \\
    \textsuperscript{2}Beijing Institute for General Artificial Intelligence \\
    \textsuperscript{3}Sun Yat-sen University \\
}


\maketitle

\begin{abstract}
In this paper, we find that the complexity of interactions encoded by a deep neural network (DNN) can explain its generalization power. We also discover that the confusing samples of a DNN, which are represented by non-generalizable interactions, are determined by its low-layer parameters. In comparison, other factors, such as high-layer parameters and network architecture, have much less impact on the composition of confusing samples. Two DNNs with different low-layer parameters usually have fully different sets of confusing samples, even though they have similar performance. This finding extends the understanding of the lottery ticket hypothesis, and well explains distinctive representation power of different DNNs.
\end{abstract}

\section{Introduction}
The explainability of deep neural networks (DNNs) has received increasing attention in recent years. Although previous studies have explained different aspects of DNNs~\citep{dziugaite2017computing, foret2020sharpness, neyshabur2015norm}, this paper focuses on a new perspective, which starts from the following two questions.

(1) Can we define and mine a set of inference patterns from a trained DNN to explain the DNN's inference socre? Can we also use these inference patterns to directly identify whether the inference/classification of a specific sample is conducted on over-fitted features? In this paper, the samples classified by over-fitted features are termed \textbf{\textit{confusing samples}}.

(2) What is the key factor that determines the composition of confusing samples of a DNN?

\begin{figure*}[t]
    \centering
    \includegraphics[width=0.99\textwidth, height=4.5cm]{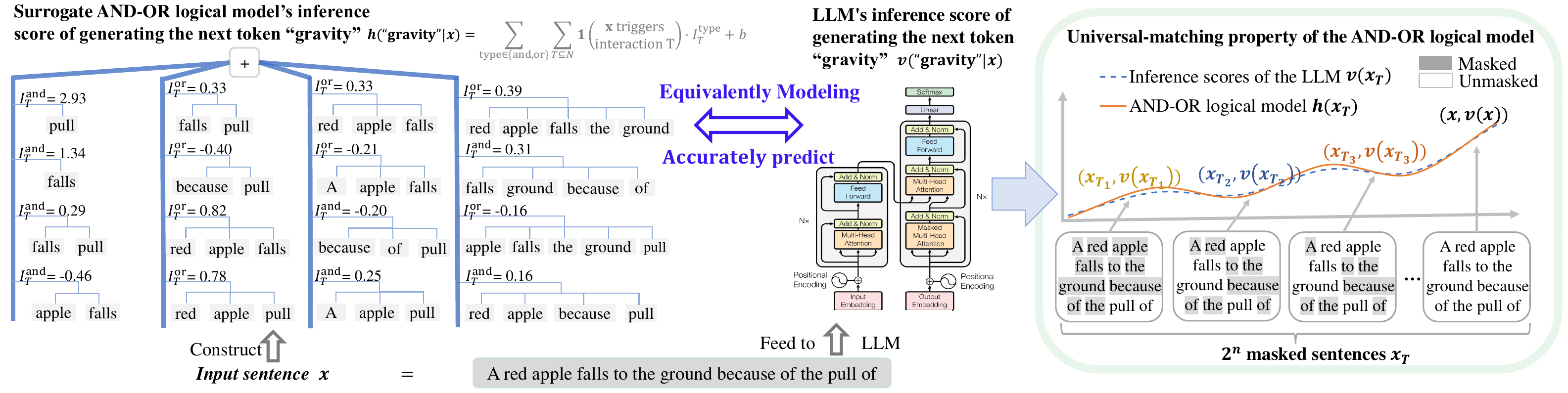}
    \vspace{-5pt}
    \caption{\citet{ren2024we} have proven that we can construct a surrogate logical model $h(\mathbf{x})$ consisting of sparse AND-OR interactions, which can universally predict the DNN's inference scores $v(\mathbf{x})$ on an exponential number of masked states of the sample $\mathbf{x}$.}
    \vspace{-5pt}
    \label{ref::fig1}
\end{figure*}

\textbf{Background.} Our research is conducted upon recent advancements in the explanation theory. \emph{I.e.,} \citet{ren2024we} have proven a series of theorems, which guarantee that given a DNN, people can use a surrogate AND-OR logical model to accurately match all network outputs on an exponential number of augmented input samples.

\textit{The above theory serves as a mathematical guarantee to let AND-OR interactions in the logical model be roughly considered as \textbf{primitive inference patterns equivalently used by the DNN for inference.}} For example, as Figure \ref{ref::fig1} shows, given an input prompt $\mathbf{x}=$``\textit{A red apple falls to the ground because of the pull of},''
the LLM generates the next token ``\textit{gravity},'' and its inference score of token generation can be faithfully explained by the interactions in the logical model, \emph{e.g.,} an AND interaction between the words in $S=\{\text{red}, \text{apple}, \text{falls}\}$ is related to ``\textit{gravity}.''

\textbf{Our research} mainly focuses on two aspects: (1) we use interactions to recognize a set of samples, to which a DNN is overfitted, and these samples are defined as the \textbf{\textit{confusing samples}}; (2) we use interactions to explore the key factor that determines the composition of confusing samples.

\textbullet \ \textbf{Using the complexity of interactions to recognize confusing samples.} Since AND-OR interactions have been proven to effectively explain varying inference scores of a DNN, interactions have been widely used as primitive inference patterns to analyze the generalization power of a DNN~\citep{zhou2024explaining, deng2022discovering}.

\begin{figure}[t]
    \centering
    \includegraphics[width=0.48\textwidth, height=2.8cm]{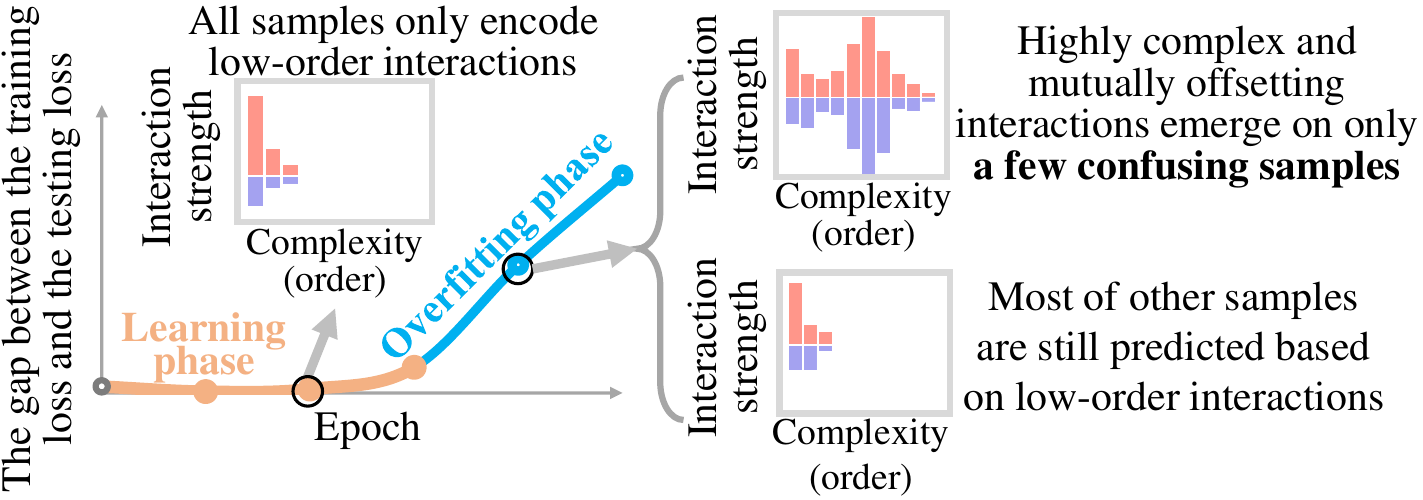}
    \vspace{-10pt}
    \caption{Complex and mutually offsetting interactions emerge only on a few samples (confusing samples) in the overfitting phase.}
    \label{ref::fig2}
\end{figure}

In this study, we use such interactions to recognize a set of \textbf{\textit{confusing samples}},  and confusing samples are defined as those whose classification/inference is conducted on non-generalizable interactions. (1) First, we find that the emergence of highly complex and mutually offsetting\footnote[1]{The complexity and the mutually offsetting are two typical properties for interactions. As Figure \ref{ref::fig1} shows, different interactions have different numerical effects. Positive/negative interactions effects push the output of the logical model towards/away from the meaning of ``gravity.'' When the DNN is over-fitted, effects of many interactions mutually offset. } interactions is the internal mechanism for the overfitting of a DNN, because such interactions are less likely to be generalized to testing samples\footnote[2]{
An interaction is considered generalizable to testing samples if it frequently appears in both training samples and testing samples and consistently pushes the DNN towards the same category.}. (2) Second, as Figure~\ref{ref::fig2} shows, when we keep training a well-trained DNN towards overfitting, the major change of a DNN is the emergence of highly complex and mutually offsetting\footnote[3]{\label{footnote-two-phase}Most high-order interactions have mutually offsetting effects, which were partially explained as noise patterns in the DNN~\cite{zhang2024two}.} interactions, but such interaction only emerge on a few samples (\emph{i.e.,} confusing samples). In comparison, the DNN's inference on other samples are still conducted on simple interactions. This explains the the DNN's overfitting towards specific samples.

\textbullet \ \textbf{Exploring the key factor that determines confusing samples.} We find that different DNNs usually have fully different sets of confusing samples, even though they have similar performance. More crucially, \textbf{we find that the randomness of parameters in low layers is the key factor that determines the composition of confusing samples of a DNN.} In comparison, other factors, such as the high-layer parameters and the network architecture, have much less impact. Specifically, if two DNNs have two different sets of low-layer parameters, then the two DNNs will have completely different sets of confusing samples, even when they have same architecture and are trained on the same dataset.

\footnotetext[4]{\label{footnote-input-variable}In image classification, the input variables can be set as different patches of an image. In language generation, the embedding vector of a token can be considered as an input variable.}

The contributions of this study can be summarized as follows. (1)
We have conducted various experiments to verify that learning complex and mutually offsetting\textsuperscript{\ref{footnote-two-phase}} interactions explains the internal mechanism for a DNN's non-generalizable representations. (2) We have discovered a counter-intuitive phenomenon that different DNNs have fully different sets of confusing samples. (3) We find that it is the randomness of a DNN's parameters in low layers that determines the composition of confusing samples of the DNN. In comparison, other factors have much less impact on the composition of confusing samples.

\section{Defining confusing samples with non-generalizable inference patterns}
\subsection{Preliminaries: extracting interactions as inference patterns used by a DNN for inference}
To explain the generalization power of a DNN using the inference patterns encoded by the DNN, the key point is how to guarantee the explained inference patterns objectively reflect the true information-processing mechanisms in the DNN. Fortunately, recent advancements in explainable AI theory~\citep{li2023does, ren2023defining, ren2024we} have discovered and proven a theoretically guaranteed faithful method to define and extract inference patterns of a DNN.

Given a DNN $v$ and an input sample \(\mathbf{x} = [x_1, x_2, \ldots, x_n]^T\) with \( n \) input variables\textsuperscript{\ref{footnote-input-variable}}, indexed by \( N = \{1, 2, \ldots, n\} \). Let \( v(\mathbf{x}) \in \mathbb{R} \) denote a scalar output of the DNN, \emph{e.g.,} the widely-used scalar classification confidence in multi-category classification~\citep{deng2022discovering}, as follows.
{
\small
\begin{align}
    v(\mathbf{x}) = \log \frac{p(y=y^{*}|\mathbf{x})}{1-p(y=y^{*}|\mathbf{x})},
\end{align}
}
where \( p(y = y^{*} | \mathbf{x}) \) represents the probability of classifying the input sample $\mathbf{x}$ to the ground-truth label $y^*$.

The universal-matching property in Theorem \ref{theorem:match}
guarantees that for each DNN $v$ and an sample $\mathbf{x}$,
we can construct a logical model $h(\mathbf{x})$ based on AND-OR interaction logics to faithfully predict all varying outputs $v(\mathbf{x})$ of the DNN on all randomly masked states $\mathbf{x}_\text{mask}$ of the sample $\mathbf{x}$.
{
\small
\begin{equation}\label{ref::universal-match}
    \begin{split}
        h(\mathbf{x}_\text{mask}) \overset{\text{def}}{=}
    \sum_{T \in \Omega_{\text{and}}}
    \underbrace{I^{\text{and}}_T \cdot \mathbb{1}(\substack{\mathbf{x}_\text{mask}\  {\scriptstyle\text{triggers AND}} \\{\scriptstyle\text{relation between}} \ T})}_{\text{an AND interaction}} +\\
    \sum_{T \in \Omega_{\text{or}}}
    \underbrace{I^{\text{or}}_T \cdot \mathbb{1}(\substack{\mathbf{x}_\text{mask}\  {\scriptstyle\text{triggers OR}} \\ {\scriptstyle\text{relation between}} \ T})}_{\text{an OR interaction}} + \ b.
    \end{split}
\end{equation}
}
\textbullet \ The binary trigger function \( \mathbb{1} \left( \substack{\mathbf{x}_\text{mask} \ {\scriptstyle \text{triggers AND}} \\ {\scriptstyle \text{relation between}} \ T} \right) \in \{0, 1\} \) represents an AND interaction among the input variables in the set \( T \). It returns \( 1 \) if all variables in \( T \) are present (not masked) in \( \mathbf{x}_\text{mask} \); otherwise, it returns \( 0 \). The scalar weight $I^{\text{and}}_T$ represents the numerical effect of the AND interaction $T$. $b=v(\emptyset)$ represents the output of the DNN when we mask all input variables in $\mathbf{x}$.

\textbullet \ The binary trigger function \( \mathbb{1} \left( \substack{\mathbf{x}_{\text{mask}} \ {\scriptstyle \text{triggers OR}} \\ {\scriptstyle \text{relation between}} \ T} \right) \in \{0, 1\} \) represents an OR interaction among input variables in the set \( T \). It returns \( 1 \) whenever any variable in \( T \) appears (not masked) in \( \mathbf{x}_\text{mask} \); otherwise, it returns \( 0 \). The scalar weight $I^{\text{or}}_T$ represents the effect of the OR interaction $T$.

\begin{theorem}[\textbf{Universal matching property}, proven in \cite{chen2024defining}]\label{theorem:match}
Given a DNN $v$ and an input sample $\mathbf{x}$, if the scalar weights $I^{\text{and}}_T$  and $I^{\text{or}}_T$ in the logical model are set as $\forall T\subseteq N, I^{\text{and}}_T = \sum\nolimits_{L \subseteq T} (-1)^{\vert T \vert - \vert L \vert} u^{\text{and}}_L$ and $I^{\text{or}}_T = - \ \sum\nolimits_{L \subseteq T} (-1)^{\vert T \vert - \vert L \vert} u^{\text{or}}_{N \backslash L}$, subject to $\forall L\subseteq N, u^{\text{and}}_L+u^{\text{or}}_L=v(\mathbf{x}_L)$, we have,
\vspace{-1pt}
\begin{equation}
    \forall S \subseteq N, \quad h(\mathbf{x}_S) = v(\mathbf{x}_S),
\end{equation}
where $\mathbf{x}_S$ represents a masked input sample only containing input variables in $S$. All other variables in $N \setminus S$ are masked\footnote[5]{
Masking an input variable in $S$ is conducted by replacing this variable with a baseline value. The baseline value is usually set to the average value of this input variable over different input samples~\citep{dabkowski2017real}.}.
\end{theorem}
\vspace{-2pt}
The universal matching property in Theorem \ref{theorem:match} shows that the logical model can always accurately predict the network outputs $v(\mathbf{x}_{\text{mask}})$, when we augment the input sample $\mathbf{x}$ by enumerating all its $2^n$ masked states. \textit{This is a powerful theorem, which guarantees that we can roughly consider the AND-OR interactions in the logical model as the \textbf{primitive inference patterns} equivalently used by the DNN.}

\textbullet \ \textbf{Sparsity property.}
Another issue with interactions is the conciseness of the interaction-based explanation. Theoretically, the logical model may contain at most $2^n$ AND interactions in $\Omega_{\text{and}}$ and $2^n$ OR interactions in $\Omega_{\text{or}}$. However, \citet{ren2024we} have proven that a well-trained DNN usually encodes only \( \Omega_{\text{and}} = \mathcal{O}(n^{\kappa} / \tau) \ll 2^n \) salient interactions,
whose absolute effects are greater than the threshold $\tau$. All other interactions have negligible effects. The salient interactions are sparse, as the empirical range of \( \kappa \) is between \( [0.9, 1.2] \). Appendix \ref{sec:apdx-condition-for-sparsity} provides conditions for the sparsity of AND-OR interactions.

\citet{zhou2023explaining} have proposed to set $\forall L\subseteq N, u^{\text{and}}_L = v(\mathbf{x}_L) +  \gamma_S$ and $u^{\text{or}}_L = v(\mathbf{x}_L) - \gamma_S$, and use a LASSO-like loss to train the parameter \(\gamma_S\) towards the sparsest interactions. Please see Appendix \ref{sec:apdx-optimize-pq} for details.

\textbf{Order of an interaction.} The order of an interaction $S$ reflects the complexity of the interaction, and is defined as the number of input variables in $S \subseteq N$, \emph{i.e.,} {\small$\text{\textit{order}}(S) = \vert S \vert$}.

\footnotetext[6]{\label{ref::salient-interactions}Due to the sparsity property of interactions, we follow \citep{ren2024towards} to define salient interactions as those with absolute effects greater than $\tau = 0.02 \mathbb{E}_{\mathbf{x}}[\vert v(\mathbf{x}) - v(\mathbf{x}_{\emptyset})\vert]$, and we only consider these salient interactions in later analysis.}
\begin{figure*}[ht]
    \centering
    \begin{minipage}[b]{0.3\textwidth}
        \centering
        \includegraphics[width=\textwidth, height=4cm]{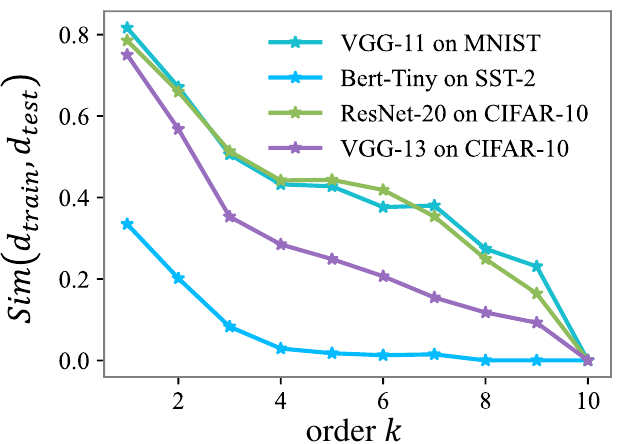}
        \vspace{-15pt}
        \caption{Jaccard similarity between interactions\textsuperscript{\ref{ref::salient-interactions}} extracted from training samples and those extracted from testing samples. Low Jaccard similarity of high-order interactions indicate the weak generalization power of high-order interactions.}
        \label{img-similarity}
    \end{minipage}%
    \hfill
    \begin{minipage}[b]{0.68\textwidth}
        \centering
        \includegraphics[width=\textwidth, height=6cm]{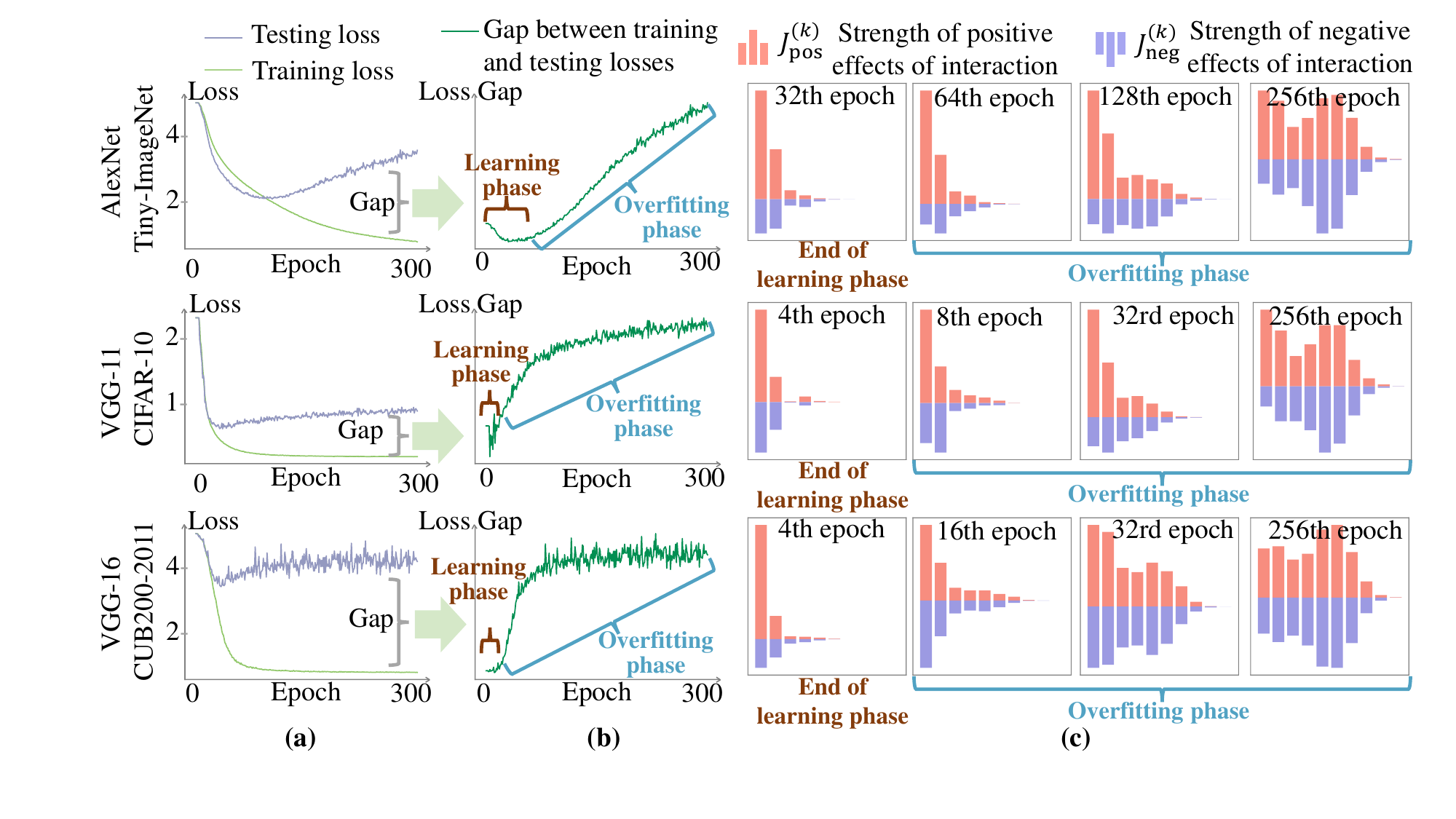}
        \vspace{-15pt}
        \caption{(a) Curves of the training loss and testing loss during the training process. (b) The loss gap between the training loss and the testing loss during the training process. (c) Distribution of interactions\textsuperscript{\ref{ref::salient-interactions}} over different orders at the end of the learning phase and during the overfitting phase. We averaged the distributions extracted from different samples. Complex and mutually offsetting interactions emerge in the over-fitting phase}
    \label{img-two-phase}
    \end{minipage}
\end{figure*}

\subsection{Connection between the interaction complexity and the generalization power}\label{ref::sec3.2}
According to Theorem~\ref{theorem:match}, the output of a DNN can be disentangled into the sum of effects of different interactions in the logical model.
Therefore, the overall generalization power of the DNN can be explained as the collective effect of the generalization power of these interactions.

Therefore, in this subsection, we conduct experiments to verify two hypothesis about the relationship between the interaction complexity and generalization power.

\textbf{How to define generalization power of interactions.} Definition~\ref{definition::similarity} is the most typical definition for the generalization power of interactions. It is shows that if an interaction frequently appears in both training samples and testing samples and consistently pushes the DNN towards the same category,  then this interaction can be considered to be generalized to the testing samples. Otherwsie, this interaction is non-generalizable.

\begin{definition}\label{definition::similarity}
    Given \( m_{\text{and}} \) AND interactions in the set \( \Phi^{\text{and}} \) and \( m_{\text{or}} \) OR interactions in the set \( \Phi^{\text{or}} \), we define the generalization power of these interactions as the Jaccard similarity $\text{Sim}(d^{\text{train}}, d^{\text{test}})$ between the distribution\footnote[7]{\label{footnote::similarity}
    We vectorize interactions extracted from a sample $\mathbf{x}$ as $\mathbf{I}(\mathbf{x})= [I^{\text{and}}_{S_1},I^{\text{and}}_{S_2},...,I^{\text{and}}_{S_{m_{\text{and}}}}, I^{\text{or}}_{S_1},I^{\text{or}}_{S_2},...,I^{\text{or}}_{S_{m_{\text{or}}}}]^T$, and compute the average effect over different training samples $\mathbf{I}_{\text{train}}=\mathbb{E}_{\mathbf{x}\in \textrm{train set}}[\mathbf{I}(\mathbf{x})]$. We follow \cite{zhou2024explaining} to separate the positive effects and negative effects and construct the vector with non-negative elements $d^{\text{train}}=\left[(\max(\mathbf{I}_{\text{train}}, \mathbf{0}))^T, (\max(-\mathbf{I}_{\text{train}}, \mathbf{0}))^T\right]^T \in \mathbb{R}^{2m_{\text{and}} + 2m_{\text{or}}}$ to represent the distribution of average effects over different interactions on training samples.
    Similarly, the vector with non-negative elements $d^{\text{test}} \in \mathbb{R}^{2m_{\text{and}} + 2m_{\text{or}}}$ represents the distribution of average effects over different testing samples.} of interactions on the training samples \( d^{\text{train}} \in \mathbb{R}^{2m_{\text{and}} + 2m_{\text{or}}} \) and the distribution\textsuperscript{\ref{footnote::similarity}} of interactions on the testing samples \( d^{\text{test}} \in \mathbb{R}^{2m_{\text{and}} + 2m_{\text{or}}} \), \emph{i.e.,} {$\text{Sim}(d^{\text{train}}, d^{\text{test}}) = \frac{\| \min(d^{\text{train}}, d^{\text{test}}) \|_1} {\| \max(d^{\text{train}}, d^{\text{test}}) \|_1}$}, where \(\Vert\cdot\Vert_1\) represents the L1-norm.
\end{definition}
\vspace{1pt}
\begin{hypothesis}\label{ref::hypothesis1}
    High-order (complex) interactions have weaker generalization power than low-order (simple) interactions.
\end{hypothesis}
\textbf{Verifying the above hypothesis about the low generalization power of high-order interactions.} This hypothesis is inspired by empirical findings in~\citep{zhou2024explaining}, which evaluates the generalization power of all interactions of each order. For each $k$-th order, there are a total of $\tbinom{n}{k}$ AND interactions and $\tbinom{n}{k}$ OR interactions belonging to this order, where $\tbinom{n}{k}$ represents the combination number of selecting $k$ variables from $n$ variables.
According to Definition~\ref{definition::similarity}, Hypothesis~\ref{ref::hypothesis1} suggests that low-order interactions usually show a higher Jaccard similarity $\text{Sim}^{(k)} \overset{\text{def}}{=}\text{Sim}(d^{\text{train}}, d^{\text{test}})_{k\text{-th order}}$ than high-order interactions. In other words, low-order interactions in training samples can be better generalized to testing samples than high-order interactions, which aligns with common intuition.

To verify Hypothesis~\ref{ref::hypothesis1}, we trained VGG-13~\citep{simonyan2014very} on the CIFAR-10 dataset~\citep{krizhevsky2009learning}, VGG-11~\citep{simonyan2014very} on the MNIST dataset~\citep{lecun1998gradient}, ResNet-20~\citep{he2016deep} on the CIFAR-10 dataset, and the Bert-Tiny model~\citep{devlin2018bert} on the SST-2 dataset~\citep{socher2013recursive} to test the generalization power of interactions\footnote[8]{\label{ref::input-varibale}
Following \citep{li2023does}, for the image classification task, we used the intermediate features of different image patches as input variables. For natural language process tasks, we used the embeddings of different input tokens as input variables. Please see Appendix \ref{sec:players} for detail setting. } of different orders in these DNNs. Figure \ref{img-similarity}
shows the generalization power of interactions of each order. The generalization power decreased as the complexity of the interactions increased, which verified Hypothesis~\ref{ref::hypothesis1}.

\textbf{Verifying the hypothesis that complex and mutually offsetting interactions emerge in the over-fitting phase.}
As Figure \ref{img-two-phase} shows, people can use the gap between the testing loss and the training loss to roughly divide the entire training process of a DNN into (1) the learning phase (where the gap is always small) and (2) the overfitting phase (where the gap begins to widen).
\citet{ren2024towards} have discovered two-phase dynamics of interactions during the DNN's training process.
Based on this, we further propose Hypothesis \ref{ref::hypothesis} to clarify the specific distribution of interactions newly emerged in the overfitting phase.

\begin{hypothesis}\label{ref::hypothesis}
The DNN mainly encodes low-order (simple) interactions on almost all training samples during the learning phase. The DNN begins to gradually encode interactions of increasing orders with mutually offsetting effects on a specific set of samples, but not all samples, during the overfitting phase.
\end{hypothesis}
To verify this hypothesis, we conducted experiments to visualize the distribution of interactions over different orders.
As Figure \ref{img-two-phase} shows, to visualize the distribution of interactions, we quantified the strength of positive interaction effects of each $k$-th order, and the strength of negative interaction effects of each $k$-th order as follows.
{
\small
\begin{equation}\label{ref::strength}
    \begin{split}
        J_{\text{pos}}^{(k)} = \sum_{\text{type} \in \{\text{and, or}\}} \sum_{|S| = k} \max(I^{\text{type}}_S, 0),\\
J_{\text{neg}}^{(k)} = \sum_{\text{type} \in \{\text{and, or}\}} \sum_{|S| = k} |\min(I^{\text{type}}_S, 0)|.
    \end{split}
\end{equation}
}

We trained AlexNet~\citep{krizhevsky2012imagenet} on the Tiny-ImageNet~\cite{tiny-imagenet} dataset, VGG-11 on the CIFAR-10 dataset, and VGG-16 on the CUB200-2011~\citep{wah2011caltech} dataset (bird images were cropped from the background).

Figure \ref{img-two-phase} shows that the DNN
usually only encoded low-order (simple) interactions at the end of the
learning phase, and began to gradually learn high-order (complex) and mutually offsetting interactions in the later overfitting phase. Figure \ref{img::average-samples} in later experiments shows that not all samples learned high-order interactions. This verified Hypothesis~\ref{ref::hypothesis}.

\textbf{Conclusion.} Based on the experimental validation of Hypothesis~\ref{ref::hypothesis1} and Hypothesis~\ref{ref::hypothesis} in Section \ref{ref::sec3.2}, we could conclude that high-order interactions were generally harder to generalize to testing samples than low-order interactions. High-order and mutually offsetting interactions are mainly learned during the overfitting phase. Thus, the emergence of high-order interactions could be seen as a typical sign of overfitting.

\subsection{Using interactions to define confusing samples}\label{red::sec3.3}
Based on the conclusion in Section \ref{ref::sec3.2}, we can use the emergence of high-order and mutually offsetting\textsuperscript{\ref{footnote-two-phase}} interactions to explain the overfitting of a DNN. Particularly, such interactions emerge on only a few training samples, rather than on all samples, during the overfitting process.
This enables us to define the specific set of training samples, in which high-order and mutually offsetting interactions emerge in the overfitting phase, as \textbf{\textit{confusing samples}}. In comparison, the distributions of interactions of other \textbf{\textit{easy (not confusing) samples}} do not change a lot.

\begin{figure}[t]
    \centering
    \includegraphics[width=0.48\textwidth, height=2.5cm]{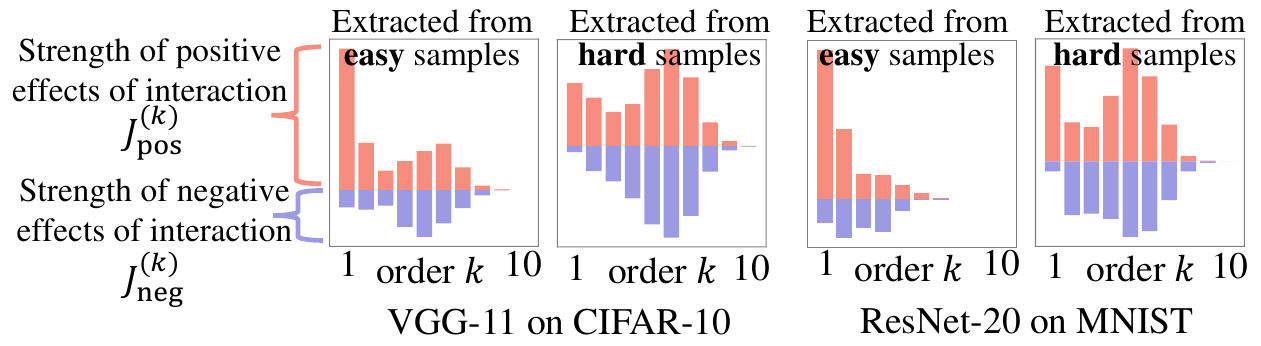}
    \vspace{-10pt}
    \caption{Distribution of interactions over different orders. We averaged the distributions extracted from different hard samples and averaged the distributions extracted from different easy samples.}
    \vspace{-5pt}
    \label{img::average-samples}
\end{figure}

\textbf{Confusing samples vs. hard samples.}
Hard samples are usually defined as samples with the highest loss~\citep{lin2017focal}. We conducted experiments to explore the relationship between hard samples and confusing samples. First, we collected a set of hard samples by selecting those with the highest loss\footnote[9]{We averaged the classification loss of each sample across different epochs during the training process, and selected the samples with the highest average loss as hard samples.} during the training process. For comparison, we randomly selected some of the remaining samples as easy samples. Then, we compared the interaction distribution of hard samples and the interaction distribution of easy samples.

Specifically, we trained VGG-11 on the CIFAR-10 dataset, and trained ResNet-20 on the MNIST dataset. Following the setting in Section~\ref{ref::sec3.2}, we visualized the interaction distribution of hard samples and the interaction distribution of easy samples. Figure \ref{img::average-samples} shows that hard samples tended to contain a large number of high-order and mutually offsetting interactions, which indicates that most hard samples were also confusing samples. In comparison, easy samples usually only encoded low-order interaction. This indicated that most easy samples were not confusing samples.

\begin{figure}[t]
    \centering
    \includegraphics[width=0.48\textwidth, height=6cm]{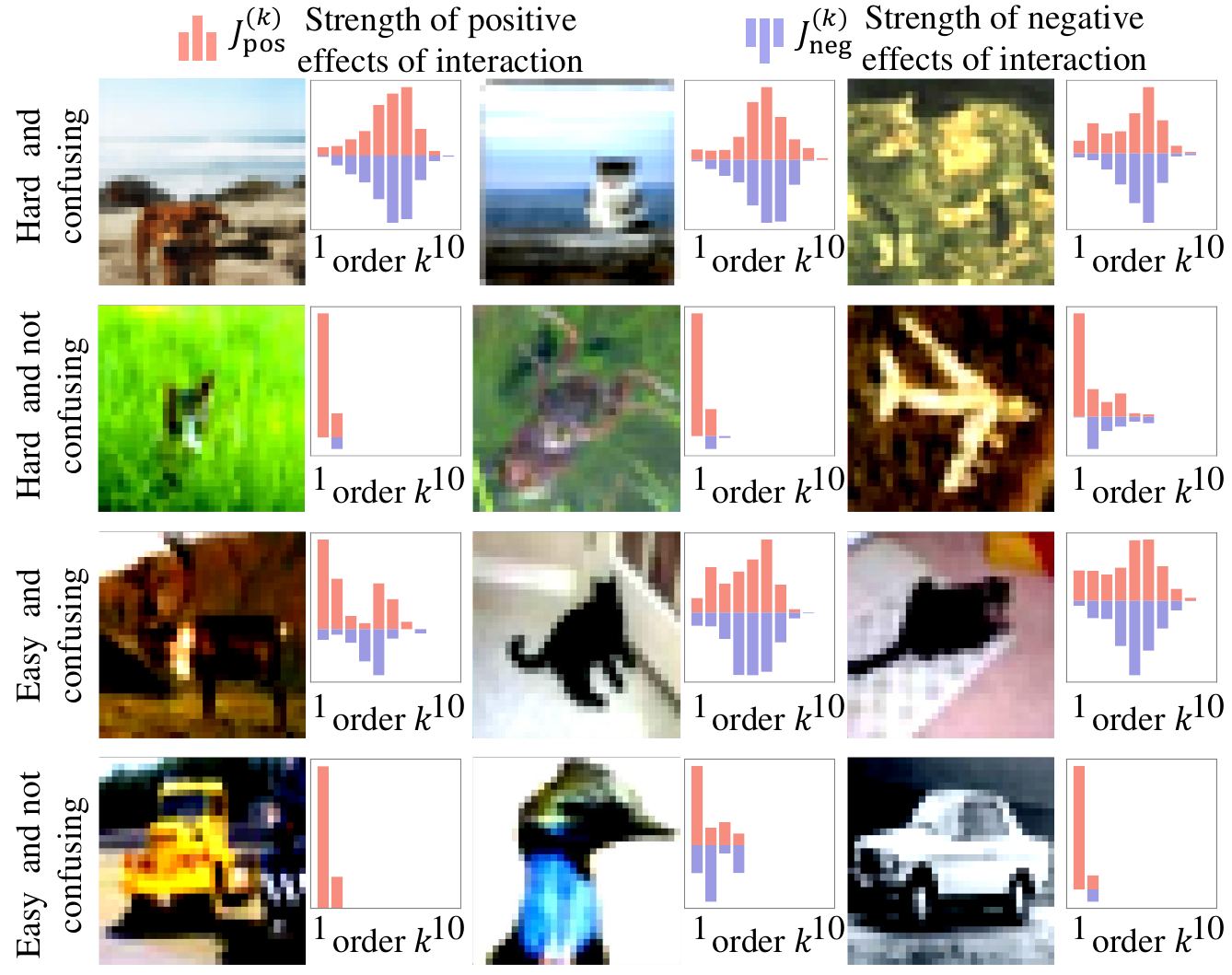}
    \vspace{-10pt}
    \caption{Confusing samples are not same as hard samples. Some hard samples are not confusing samples, and some confusing samples are not hard samples, either.}
    \vspace{-10pt}
    \label{img::specific-samples}
\end{figure}

However, as Figure~\ref{img::specific-samples} shows, although there is a considerable overlap between confusing samples and hard samples, they are not exactly the same.

\textit{First, a considerable ratio of hard samples are not confusing samples.} Hard samples can usually be categorized into two types from the perspective of interactions. (1) Most hard samples encode mutually offsetting\textsuperscript{\ref{footnote-two-phase}} interactions, which weaken the classification confidence. (2) The other type of hard samples only have a few interactions. It is the small number of interactions, not the mutually offsetting of interactions, that weakens the classification confidence. To this end, only the fist type of hard samples can be explained as confusing samples.

\textit{Second, some confusing samples are not hard samples, either.} Although the encoding of mutually offsetting interactions in confusing samples usually significantly hurts the classification confidence according to Equation (\ref{ref::universal-match}), some confusing samples may still be classified with high confidence. As Figure \ref{img::specific-samples} shows, such samples usually contain a large number of interactions, including both lots of mutually offsetting interactions and numerous non-offsetting low-order interactions.
The large interaction number can also enhance the sample's classification confidence, according to Equation (\ref{ref::universal-match}).

\textbf{The emergence of high-order interactions on confusing samples is the main phenomenon during the overfitting phase of a DNN.} Traditionally, hard samples are believed to be the primary factor that pushes a DNN towards overfitting. However, our experiments in Section~\ref{ref::sec3.2} show that confusing samples played a distinctive role that contributed to the overfitting of a DNN.

As Figure \ref{img-two-phase} shows, when the training process of a DNN entered the overfitting phase (where the gap between the testing loss and training loss began to widen), we observed that the DNN often encoded a large number of high-order and mutually offsetting interactions. Moreover, it had been proven that the output of a DNN could be disentangled as the sum of the effects of interactions. As a result, encoding these mutually offsetting and high-order interactions caused the DNN to gradually transit into the overfitting phase. In this way, these mutually offsetting and high-order interactions could be roughly considered as an explanation for the overfitting of a DNN.

\section{Exploring  the key factor that determines the confusing samples encoded by a DNN}
Currently, many engineering techniques have been proposed to enhance the generalization power of DNNs and prevent overfitting, such as improvements of the network architecture~\cite{he2016deep}, data cleaning~\cite{northcutt2021confident}, and data augmentation~\cite{shorten2019survey}.

However, despite previous studies, it is still unclear which factor determines the composition of confusing samples.
Therefore, in this study,  we conduct experiments and find that it is the randomness of low-layer parameters that determines the composition of confusing samples of a DNN. In comparison, other factors, such as the network’s architecture and the parameters in the high layers, have much less impact on the composition of confusing samples.

\subsection{Randomness of confusing samples}\label{red::sec4.1}
We find a counter-intuitive phenomenon, \emph{i.e.,} different DNNs with similar classification performance usually have fully different sets of confusing samples. This finding seems to conflict with another closely related topic, \emph{i.e.,} \textit{mining hard samples}, which considers the composition of hard samples is an intrinsic property of data distribution in a high-dimension space.
People usually assume different AI models share the same set of hard samples, and this idea has been widely used for data augmentation~\cite{shrivastava2016training, smirnov2018hard, peng2018jointly}.

However, the following phenomenon of the randomness of confusing samples challenges the above well-known common sense.
Later, this phenomenon is found to be attributed to the randomness of parameters in low layers of the DNN in Section~\ref{ref::sec4.2}.

\textbf{Phenomenon 1.} \textit{DNNs with similar classification accuracies, even those with the same architecture, usually had completely different sets of confusing samples.
}

\textbf{Metric.} Given all interactions extracted from a given sample $\mathbf{x}$, we use the average order of interactions extracted from $\mathbf{x}$, $\eta^{avg} = \sum_{k=1}^{n} (k \cdot J_{\text{pos}}^{(k)} + k \cdot J_{\text{neg}}^{(k)}) / \sum_{k=1}^{n}(J_{\text{pos}}^{(k)} + J_{\text{neg}}^{(k)})$ as a metric to roughly distinguish whether the given sample is a confusing sample. The average order of interactions is weighted by the interaction strength of each order $J_{\text{pos}}^{(k)}$ and $J_{\text{neg}}^{(k)}$ in Equation (\ref{ref::strength}). Here, we only considere the complexity (order) of interactions, and ignore the mutually offsetting of interactions, because it is rare to find high-order interactions without mutually offsetting effects in real experiments. In this way, all samples with a high value of $\eta^{avg}$ can be simply considered as confusing samples.

\begin{figure}[t]
    \centering
    \includegraphics[width=0.48\textwidth, height=4.5cm]{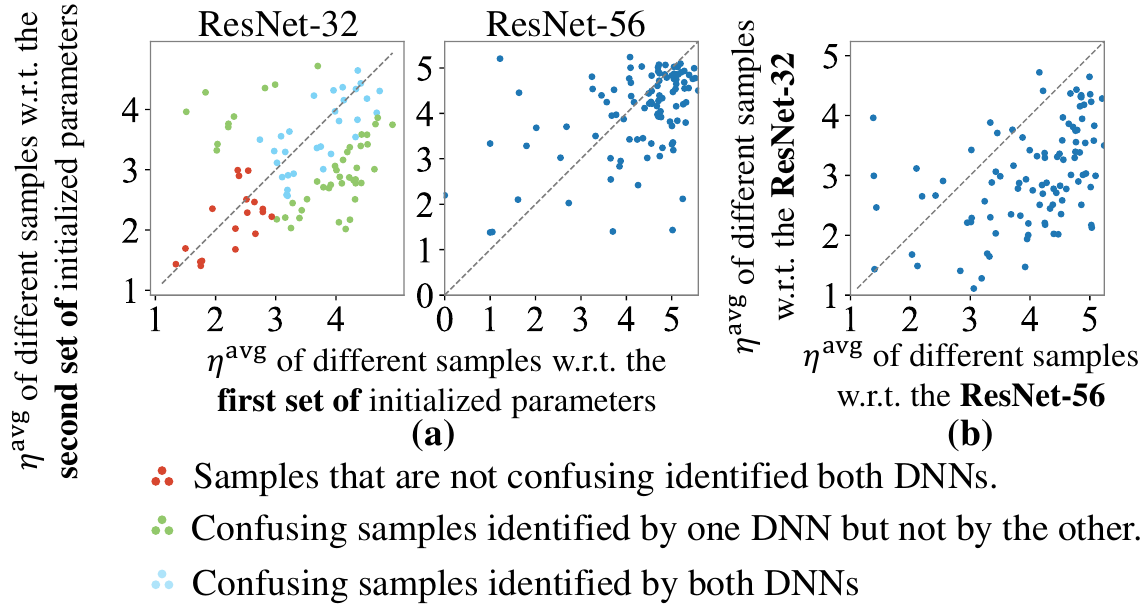}
    \vspace{-15pt}
    \caption{Comparing composition of confusing samples of two DNNs, which are trained on the same dataset. (a, b) The DNNs with the \textbf{same} architecture but \textbf{different} initialized parameters had different confusing samples. (c) The DNNs with \textbf{different} architectures (of course, different low-layer paraemeters) had different confusing samples.}
    \vspace{-10pt}
    \label{ref::fig41}
\end{figure}

We use the scatter diagram in Figure~\ref{ref::fig41} to identify whether two DNNs have similar sets of confusing samples. Each point in the figure represents a sample. The horizontal axis shows a sample's average interactions order $\eta^{avg}$ extracted from a DNN, and the vertical axis shows its average interaction order $\eta^{avg}$ extracted from the other DNN.
If the two DNNs have similar sets of confusing samples, then most points (samples) will appear near the main diagonal of the figure. Otherwise, if many confusing samples for a DNN are not confusing samples for the other DNN, then these sample would deviate from the main diagonal of the figure.

\begin{figure*}[t]
    \centering
    \includegraphics[width=0.98\textwidth, height=6cm]{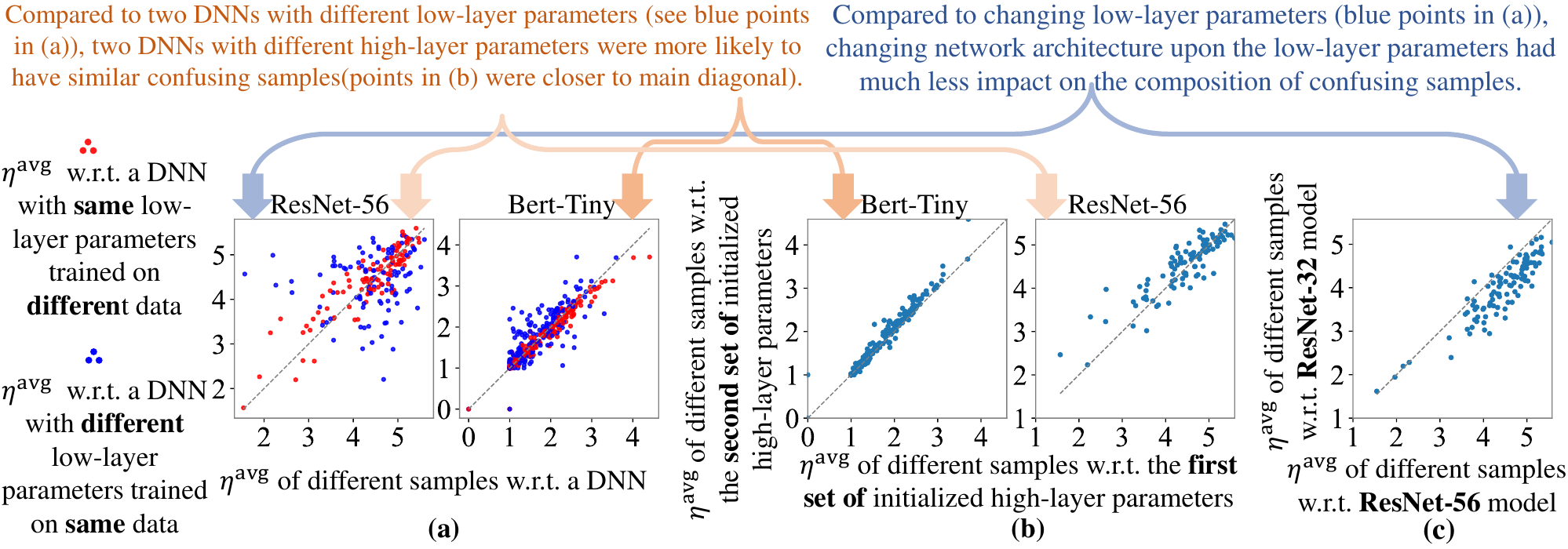}
    \vspace{-5pt}
    \caption{(a) Red points compare composition of confusing samples of two DNNs with the \textbf{same} low-layer parameters, which are trained on \textbf{different} datasets. Blue points compare composition of confusing samples of two DNNs with the \textbf{different} low-layer parameters, which are trained on the \textbf{same} datasets. (b) Comparing composition of confusing samples of two DNNs with the \textbf{same} low-layer parameters but \textbf{different} high-layer parameters, which are trained on the \textbf{same} datasets. (c) Comparing composition of confusing samples of two DNNs with the  \textbf{same} low-layer parameters but \textbf{different} architectures, which are trained on the \textbf{same} datasets.}
    \label{ref::fig42}
\end{figure*}

\textbf{Experiments.} We conducted experiments on ResNet-32 and ResNet-56 trained on the CIFAR-10 dataset. Figure \ref{ref::fig41}(a) shows that, if parameters in two DNNs were initialized differently, then the two DNNs usually had fully different sets of confusing samples, \emph{i.e.,} most samples deviated from the main diagonal of the figure. Figure \ref{ref::fig41}(b) shows that DNNs with different architectures (thereby, having different low-layer parameters) had completely different sets of confusing samples.

\textbf{Challenging the traditional view of the sample's difficulty.} The above experiments challenge the common belief that difficulty of samples in a dataset is an intrinsic property of the data itself, although the data simplicity still cannot be ignored, either\footnote[10]{For example, it's hard to imagine that handwritten digit samples will become confusing when trained alongside CIFAR images.}. In other words, previous studies~\cite{forouzesh2024differences} usually believe that different models have similar sets of hard samples.
Although confusing samples are not fully equivalent to the hard samples, our finding suggests that the simplicity of an sample is not the only factor that determines a confusing sample, especially for the DNN. Intead, later experiments in Section \ref{ref::sec4.2} will show that it is the randomness of low-layer parameters that determines the composition of confusing samples.

\subsection{Impact of parameters in low layers}\label{ref::sec4.2}
In this subsection, we conducted experiments to analyze the impact of a DNN's low-layer parameters on the composition of confusing samples.

\textbf{Impact of low-layer parameters.} The first experiment compared the confusing samples extracted from two DNNs with exactly the same architecture but different parameters in low layers. Both DNNs were initialized to have the same parameters in high layers. As the only difference between them, we set parameters in low layers to have fully different sets of values\footnote[11]{\label{ref::fix-setting}We used low-layer parameters of another two DNNs, which had the same architecture and had been well trained, to replace low-layer parameters of the current two DNNs, respectively.}. Then, the two DNNs were trained on the same dataset to ensure a fair comparison. We trained the ResNet-56 models on the CIFAR-10 dataset and trained the Bert-Tiny models on the SST-2 dataset.
Specifically, we empirically considered the first 9 convolutional layers of ResNet-56 as the low layers, and considered all the other 47 layers as the high layers. For the Bert-Tiny model, we considered the first transformer block as the low layers, and considered all layers after the first transformer block as the high layers.

Blue points in Figure \ref{ref::fig42}(a) compares two sets of confusing samples\footnote[12]{\label{ref::confusing-sample}The identification of confusing samples was conducted on testing samples, not training samples, in all experiments for a fair comparison.} extracted from two DNNs with different low-layer parameters. We found that DNNs with different low-layer parameters usually had completely different sets of confusing samples.

\textbf{Comparison with impact of training samples.} The second experiment illustrated the impact of training samples on the composition of confusing samples\textsuperscript{\ref{ref::confusing-sample}}, so that we could compare the impact of training samples with the impact of low-layer parameters in Figure \ref{ref::fig42}(a).
For implementation, we set two DNNs, which had the same architecture, to have the same low-layer parameters\footnote[13]{\label{ref::fix-a-DNN}We used low-layer parameters of another a well-trained DNN to replace low-layer parameters of the current two DNNs, so as to let the current two DNNS have the same low-layer parameters.} and the same initialized high-layer parameters. The two DNNs were trained on different datasets.
Specifically, we prepared two disjoint training sets, each containing 1000 samples sampled from the CIFAR-10 training set for two ResNet-56 models. Similarly, we randomly sampled two disjoint sets of 2000 training samples from the SST-2 training set for two Bert-Tiny models. Red points in Figure \ref{ref::fig42}(a) compare two sets of confusing samples\textsuperscript{\ref{ref::confusing-sample}} extracted from the two DNNs. We found that compared to changing low-layer parameters in the first experiment (see blue points in Figure \ref{ref::fig42}(a)), changing training samples in the second experiment (red points) was less likely to jumble up the composition of confusing samples.

\begin{figure}[t]
    \centering
    \includegraphics[width=0.48\textwidth, height=3.5cm]{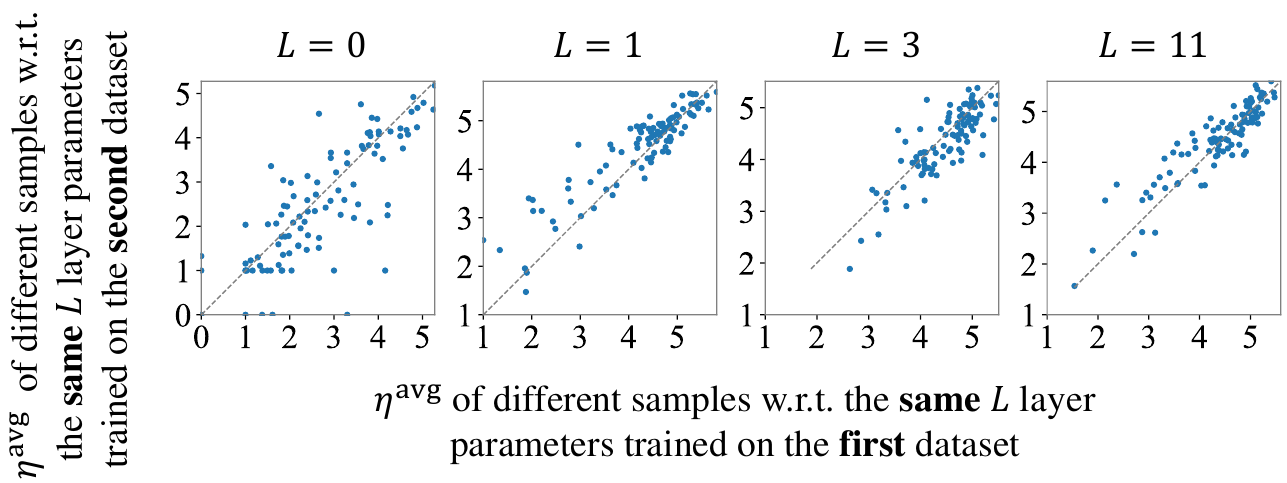}
    \vspace{-15pt}
    \caption{Comparing the composition of confusing samples of two DNNs,
    when the two DNNs had the same parameters of the first $L$ layers. The two DNNs were trained on different datasets.}
    \vspace{-5pt}
    \label{ref::fig44}
\end{figure}

\textbf{Parameters in how many layers are sufficient to determine confusing samples?}
We futher conducted experiments to explore controlling parameters in how many low layers were sufficient to determine the composition of confusing samples. Specifically, let two target DNNs with the same architecture have the same low-layer parameters, \emph{i.e.,} copying their parameters in the first $L$ layers from the same trained neural network. Then, we trained the upper layers of the two DNNs on two different sets of training samples.

We trained four different pairs of DNNs. The four pairs of DNNs shared parameters of the first 11 layers, the first 3 layers, the first 1 layer, and 0 layer (no parameter sharing), respectively. Each pair of DNNs were trained on two sets of 1000 training samples from the CIFAR-10 dataset. Figure \ref{ref::fig44} shows that sharing parameters in the first layer had already been enough to let two DNNs have similar sets of confusing samples\textsuperscript{\ref{ref::confusing-sample}}, even though the two DNNs were trained on different training samples.

\subsection{Impact of high-layer parameters and architecture}\label{ref::sec4.3}
\textbf{Impact of high-layer parameters.} We conducted an experiment to analyze the impact of parameters in high layers of a DNN. Specifically, we constructed two DNNs with exactly the same architecture, but set them to have fully different initial parameters in high layers. In comparison, the two DNNs were set with the same low-layer parameters\textsuperscript{\ref{ref::fix-a-DNN}}.

We trained such a pair of ResNet-56 models on the CIFAR-10 dataset, and trained a pair of Bert-Tiny models on the SST-2 dataset. We followed the experimental setting of high/low layers in Section~\ref{ref::sec4.2}. Figure \ref{ref::fig42}(b) shows that DNNs with different high-layer parameters still had similar sets of confusing samples\textsuperscript{\ref{ref::confusing-sample}}, which suggested that parameters in high layers had relatively little impact.

\textbf{Impact of network architecture.}
We conducted an experiment to analyze the impact of network architecture on the composition of confusing samples. Specifically, we followed experimental settings in Section~\ref{ref::sec4.2} to set two DNNs to have exactly the same parameters in low layers, but the two DNNs were constructed to have fully different architectures upon the low layers. For example, we trained ResNet-56 and ResNet-32 on the CIFAR-10 dataset,
and empirically set the first 9 layers of the two DNNs to have the same parameters\textsuperscript{\ref{ref::fix-a-DNN}}.

Figure \ref{ref::fig42}(c) shows that DNNs with different architectures also had similar set of confusing samples\textsuperscript{\ref{ref::confusing-sample}}, which indicated that network architecture had relatively weak impact.

\subsection{How to understand the randomness of low-layer parameters}
\textbf{Connection to the lottery tickect hypothesis.} Our findings extend the lottery ticket hypothesis~\cite{frankle2018lottery}. The lottery ticket hypothesis suggests that the representation of a DNN is dominated by a small set of randomly initialized parameters, which are termed \textit{winning tickets}. To this end, our experiments further showed that it was the randomness of low-layer parameters that determined the composition of confusing samples of the DNN. In comparison, other factors, such as high-layer parameters and network architecture, had much less impact.

More crucially, the low-layer parameters are usually more difficult to optimize than high-layer parameters. This means that the composition of confusing samples of a DNN is primarily determined by the initialization of low-layer parameters, regardless of how we design the architecture and parameters in high layers.

\textbf{The randomness of confusing samples reflects distinctive property for each DNN.}
The randomness of confusing samples observed in Sections~\ref{red::sec4.1} and \ref{ref::sec4.2} provides another distinctive property of confusing samples. Unlike hard samples mainly describing intrinsic nature of data distribution, the composition of confusing samples seems to be fully determined by the uncertainty (the randomness) of the low-layer parameters, without any clear pattern. This represents a distinctive property of each DNN.

\section{Conclusions}
In this paper, we have verified that learning complex and mutually offsetting interactions in a set of confusing samples explains the internal mechanism for a DNN's non-generalizable representations. Moreover, we have discovered that DNNs often have fully different sets of confusing samples. It is the randomness of low-layer parameters that determines the composition of confusing samples of the DNN. In comparison, other factors, such as high-layer parameters and network architecture, have much less impact on the composition of confusing samples.

\bibliographystyle{plainnat}
\bibliography{main}

\begin{thebibliography}{36}
\providecommand{\natexlab}[1]{#1}
\providecommand{\url}[1]{\texttt{#1}}
\expandafter\ifx\csname urlstyle\endcsname\relax
  \providecommand{\doi}[1]{doi: #1}\else
  \providecommand{\doi}{doi: \begingroup \urlstyle{rm}\Url}\fi

\bibitem[Chen et~al.(2024)Chen, Lou, Huang, and Zhang]{chen2024defining}
Lu~Chen, Siyu Lou, Benhao Huang, and Quanshi Zhang.
\newblock Defining and extracting generalizable interaction primitives from {DNN}s.
\newblock In \emph{The Twelfth International Conference on Learning Representations}, 2024.
\newblock URL \url{https://openreview.net/forum?id=OCqyFVFNeF}.

\bibitem[Dabkowski and Gal(2017)]{dabkowski2017real}
Piotr Dabkowski and Yarin Gal.
\newblock Real time image saliency for black box classifiers.
\newblock \emph{Advances in neural information processing systems}, 30, 2017.

\bibitem[Deng et~al.(2022)Deng, Ren, Zhang, and Zhang]{deng2022discovering}
Huiqi Deng, Qihan Ren, Hao Zhang, and Quanshi Zhang.
\newblock {DISCOVERING} {AND} {EXPLAINING} {THE} {REPRESENTATION} {BOTTLENECK} {OF} {DNNS}.
\newblock In \emph{International Conference on Learning Representations}, 2022.
\newblock URL \url{https://openreview.net/forum?id=iRCUlgmdfHJ}.

\bibitem[Deng et~al.(2024)Deng, Zou, Du, Chen, Feng, Yang, Li, and Zhang]{deng2024unifying}
Huiqi Deng, Na~Zou, Mengnan Du, Weifu Chen, Guocan Feng, Ziwei Yang, Zheyang Li, and Quanshi Zhang.
\newblock Unifying fourteen post-hoc attribution methods with taylor interactions.
\newblock \emph{IEEE Transactions on Pattern Analysis and Machine Intelligence}, 2024.

\bibitem[Devlin(2018)]{devlin2018bert}
Jacob Devlin.
\newblock Bert: Pre-training of deep bidirectional transformers for language understanding.
\newblock \emph{arXiv preprint arXiv:1810.04805}, 2018.

\bibitem[Dziugaite and Roy(2017)]{dziugaite2017computing}
Gintare~Karolina Dziugaite and Daniel~M Roy.
\newblock Computing nonvacuous generalization bounds for deep (stochastic) neural networks with many more parameters than training data.
\newblock \emph{arXiv preprint arXiv:1703.11008}, 2017.

\bibitem[Foret et~al.(2020)Foret, Kleiner, Mobahi, and Neyshabur]{foret2020sharpness}
Pierre Foret, Ariel Kleiner, Hossein Mobahi, and Behnam Neyshabur.
\newblock Sharpness-aware minimization for efficiently improving generalization.
\newblock \emph{arXiv preprint arXiv:2010.01412}, 2020.

\bibitem[Forouzesh and Thiran(2024)]{forouzesh2024differences}
Mahsa Forouzesh and Patrick Thiran.
\newblock Differences between hard and noisy-labeled samples: An empirical study.
\newblock In \emph{Proceedings of the 2024 SIAM International Conference on Data Mining (SDM)}, pages 91--99. SIAM, 2024.

\bibitem[Frankle and Carbin(2018)]{frankle2018lottery}
Jonathan Frankle and Michael Carbin.
\newblock The lottery ticket hypothesis: Finding sparse, trainable neural networks.
\newblock \emph{arXiv preprint arXiv:1803.03635}, 2018.

\bibitem[Harsanyi(1963)]{harsanyi1963}
John~C. Harsanyi.
\newblock A simplified bargaining model for the n-person cooperative game.
\newblock \emph{International Economic Review}, 4\penalty0 (2):\penalty0 194--220, 1963.
\newblock ISSN 00206598, 14682354.

\bibitem[He et~al.(2016)He, Zhang, Ren, and Sun]{he2016deep}
Kaiming He, Xiangyu Zhang, Shaoqing Ren, and Jian Sun.
\newblock Deep residual learning for image recognition.
\newblock In \emph{Proceedings of the IEEE conference on computer vision and pattern recognition}, pages 770--778, 2016.

\bibitem[Krizhevsky et~al.(2009)Krizhevsky, Hinton, et~al.]{krizhevsky2009learning}
Alex Krizhevsky, Geoffrey Hinton, et~al.
\newblock Learning multiple layers of features from tiny images.
\newblock 2009.

\bibitem[Krizhevsky et~al.(2012)Krizhevsky, Sutskever, and Hinton]{krizhevsky2012imagenet}
Alex Krizhevsky, Ilya Sutskever, and Geoffrey~E Hinton.
\newblock Imagenet classification with deep convolutional neural networks.
\newblock \emph{Advances in neural information processing systems}, 25, 2012.

\bibitem[LeCun et~al.(1998)LeCun, Bottou, Bengio, and Haffner]{lecun1998gradient}
Yann LeCun, L{\'e}on Bottou, Yoshua Bengio, and Patrick Haffner.
\newblock Gradient-based learning applied to document recognition.
\newblock \emph{Proceedings of the IEEE}, 86\penalty0 (11):\penalty0 2278--2324, 1998.

\bibitem[Li and Zhang(2023{\natexlab{a}})]{li2023defining}
Mingjie Li and Quanshi Zhang.
\newblock Defining and quantifying and-or interactions for faithful and concise explanation of dnns.
\newblock \emph{arXiv preprint arXiv:2304.13312}, 2023{\natexlab{a}}.

\bibitem[Li and Zhang(2023{\natexlab{b}})]{li2023does}
Mingjie Li and Quanshi Zhang.
\newblock Does a neural network really encode symbolic concepts?
\newblock In \emph{International conference on machine learning}, pages 20452--20469. PMLR, 2023{\natexlab{b}}.

\bibitem[Lin(2017)]{lin2017focal}
T~Lin.
\newblock Focal loss for dense object detection.
\newblock \emph{arXiv preprint arXiv:1708.02002}, 2017.

\bibitem[Madry(2017)]{madry2017towards}
Aleksander Madry.
\newblock Towards deep learning models resistant to adversarial attacks.
\newblock \emph{arXiv preprint arXiv:1706.06083}, 2017.

\bibitem[mnmoustafa(2017)]{tiny-imagenet}
Mohammed~Ali mnmoustafa.
\newblock Tiny imagenet, 2017.
\newblock URL \url{https://kaggle.com/competitions/tiny-imagenet}.

\bibitem[Neyshabur et~al.(2015)Neyshabur, Tomioka, and Srebro]{neyshabur2015norm}
Behnam Neyshabur, Ryota Tomioka, and Nathan Srebro.
\newblock Norm-based capacity control in neural networks.
\newblock In \emph{Conference on learning theory}, pages 1376--1401. PMLR, 2015.

\bibitem[Northcutt et~al.(2021)Northcutt, Jiang, and Chuang]{northcutt2021confident}
Curtis Northcutt, Lu~Jiang, and Isaac Chuang.
\newblock Confident learning: Estimating uncertainty in dataset labels.
\newblock \emph{Journal of Artificial Intelligence Research}, 70:\penalty0 1373--1411, 2021.

\bibitem[Peng et~al.(2018)Peng, Tang, Yang, Feris, and Metaxas]{peng2018jointly}
Xi~Peng, Zhiqiang Tang, Fei Yang, Rogerio~S Feris, and Dimitris Metaxas.
\newblock Jointly optimize data augmentation and network training: Adversarial data augmentation in human pose estimation.
\newblock In \emph{Proceedings of the IEEE conference on computer vision and pattern recognition}, pages 2226--2234, 2018.

\bibitem[Ren et~al.(2023)Ren, Li, Chen, Deng, and Zhang]{ren2023defining}
Jie Ren, Mingjie Li, Qirui Chen, Huiqi Deng, and Quanshi Zhang.
\newblock Defining and quantifying the emergence of sparse concepts in dnns.
\newblock In \emph{Proceedings of the IEEE/CVF conference on computer vision and pattern recognition}, pages 20280--20289, 2023.

\bibitem[Ren et~al.(2024{\natexlab{a}})Ren, Gao, Shen, and Zhang]{ren2024we}
Qihan Ren, Jiayang Gao, Wen Shen, and Quanshi Zhang.
\newblock Where we have arrived in proving the emergence of sparse interaction primitives in ai models.
\newblock In \emph{The Twelfth International Conference on Learning Representations}, 2024{\natexlab{a}}.

\bibitem[Ren et~al.(2024{\natexlab{b}})Ren, Zhang, Xu, Xin, Liu, and Zhang]{ren2024towards}
Qihan Ren, Junpeng Zhang, Yang Xu, Yue Xin, Dongrui Liu, and Quanshi Zhang.
\newblock Towards the dynamics of a {DNN} learning symbolic interactions.
\newblock In \emph{The Thirty-eighth Annual Conference on Neural Information Processing Systems}, 2024{\natexlab{b}}.
\newblock URL \url{https://openreview.net/forum?id=dIHXwKjXRE}.

\bibitem[Robbins and Monro(1951)]{robbins1951stochastic}
Herbert Robbins and Sutton Monro.
\newblock A stochastic approximation method.
\newblock \emph{The annals of mathematical statistics}, pages 400--407, 1951.

\bibitem[Shorten and Khoshgoftaar(2019)]{shorten2019survey}
Connor Shorten and Taghi~M Khoshgoftaar.
\newblock A survey on image data augmentation for deep learning.
\newblock \emph{Journal of big data}, 6\penalty0 (1):\penalty0 1--48, 2019.

\bibitem[Shrivastava et~al.(2016)Shrivastava, Gupta, and Girshick]{shrivastava2016training}
Abhinav Shrivastava, Abhinav Gupta, and Ross Girshick.
\newblock Training region-based object detectors with online hard example mining.
\newblock In \emph{Proceedings of the IEEE conference on computer vision and pattern recognition}, pages 761--769, 2016.

\bibitem[Simonyan(2014)]{simonyan2014very}
Karen Simonyan.
\newblock Very deep convolutional networks for large-scale image recognition.
\newblock \emph{arXiv preprint arXiv:1409.1556}, 2014.

\bibitem[Smirnov et~al.(2018)Smirnov, Melnikov, Oleinik, Ivanova, Kalinovskiy, and Luckyanets]{smirnov2018hard}
Evgeny Smirnov, Aleksandr Melnikov, Andrei Oleinik, Elizaveta Ivanova, Ilya Kalinovskiy, and Eugene Luckyanets.
\newblock Hard example mining with auxiliary embeddings.
\newblock In \emph{Proceedings of the IEEE Conference on Computer Vision and Pattern Recognition Workshops}, pages 37--46, 2018.

\bibitem[Socher et~al.(2013)Socher, Perelygin, Wu, Chuang, Manning, Ng, and Potts]{socher2013recursive}
Richard Socher, Alex Perelygin, Jean Wu, Jason Chuang, Christopher~D Manning, Andrew~Y Ng, and Christopher Potts.
\newblock Recursive deep models for semantic compositionality over a sentiment treebank.
\newblock In \emph{Proceedings of the 2013 conference on empirical methods in natural language processing}, pages 1631--1642, 2013.

\bibitem[Sundararajan et~al.(2020)Sundararajan, Dhamdhere, and Agarwal]{sundararajan2020shapley}
Mukund Sundararajan, Kedar Dhamdhere, and Ashish Agarwal.
\newblock The shapley taylor interaction index.
\newblock In \emph{International Conference on Machine Learning}, pages 9259--9268. PMLR, 2020.

\bibitem[Wah et~al.(2011)Wah, Branson, Welinder, Perona, and Belongie]{wah2011caltech}
Catherine Wah, Steve Branson, Peter Welinder, Pietro Perona, and Serge Belongie.
\newblock The caltech-ucsd birds-200-2011 dataset.
\newblock 2011.

\bibitem[Zhang et~al.(2024)Zhang, Li, Lin, and Zhang]{zhang2024two}
Junpeng Zhang, Qing Li, Liang Lin, and Quanshi Zhang.
\newblock Two-phase dynamics of interactions explains the starting point of a dnn learning over-fitted features.
\newblock \emph{arXiv preprint arXiv:2405.10262}, 2024.

\bibitem[Zhou et~al.(2023)Zhou, Tang, Li, Zhang, Liu, and Zhang]{zhou2023explaining}
Huilin Zhou, Huijie Tang, Mingjie Li, Hao Zhang, Zhenyu Liu, and Quanshi Zhang.
\newblock Explaining how a neural network play the go game and let people learn.
\newblock \emph{arXiv preprint arXiv:2310.09838}, 2023.

\bibitem[Zhou et~al.(2024)Zhou, Zhang, Deng, Liu, Shen, Chan, and Zhang]{zhou2024explaining}
Huilin Zhou, Hao Zhang, Huiqi Deng, Dongrui Liu, Wen Shen, Shih-Han Chan, and Quanshi Zhang.
\newblock Explaining generalization power of a dnn using interactive concepts.
\newblock In \emph{Proceedings of the AAAI Conference on Artificial Intelligence}, volume~38, pages 17105--17113, 2024.

\end{thebibliography}

\newpage
\appendix
\onecolumn

\section{Related work}

The explainability of deep neural networks (DNNs) has received increasing attention in recent years. However, there has long been a pessimistic view regarding the possibility of faithfully explaining DNNs's inference logics~\citep{dziugaite2017computing, foret2020sharpness, neyshabur2015norm}. Fortunately, recent advancements in interaction-based explanations, as surveyed by \cite{ren2024we}, have made the first attempt to tackle the mathematical feasibility of explaining a DNN's inference logics using a small number of inference patterns. Specifically, (1) \citet{ren2023defining} discovered and \cite{ren2024we} proved that there exists an AND-OR logical model, which contains only a small number of interactions, can faithfully explain the inference logics of DNNs, regardless of how the input samples are masked. (2) \citet{zhou2024explaining} used the complexity of interactions to explain the generalization power of DNNs. (3) \citet{deng2024unifying} demonstrated that fourteen attribution methods can all be explained as a reallocation of interaction effects.

In this way, compared to previous studies, this paper provides further insights into the underlying factors contributing to the overfitting of DNNs and identifies the key factor that determines the composition of confusing samples in DNNs. The lottery ticket hypothesis\cite{frankle2018lottery} suggests that a DNN's representation is largely influenced by a small subset of randomly initialized parameters, known as winning tickets. Building on this, our experiments showed that the low-layer parameters of a DNN are the primary determinant of the composition of confusing samples. In contrast, other factors, such as high-layer parameters and network architecture, have significantly less impact.

\section{Properties of the AND interaction}
\label{sec:apdx-property-harsanyi}

The Harsanyi interaction \cite{harsanyi1963} (referred to as the AND interaction in this work) has been a conventional metric for measureing the effect of the AND relationship that a DNN encodes among input variables. In this section, we introduce several desirable axioms that the AND interaction {\small$I^{\text{and}}_T$} adheres to. These properties further underscore the reliability of using AND interactions to explain the inference score of a DNN.

(1) \textit{Efficiency axiom} (proven by \cite{harsanyi1963}). The output score of a model can be decomposed into interaction effects of different patterns, \emph{i.e.} {\small $v(\mathbf{x})=\sum_{T\subseteq N}I^{\text{and}}_T$}.

(2) \textit{Linearity axiom}. If we merge output scores of two models $v_1$ and $v_2$ as the output of model $v$, \emph{i.e.} {\small $\forall S\subseteq N,~ v(\mathbf{x}_S)=v_1(\mathbf{x}_S)+v_2(\mathbf{x}_S)$}, then their interaction effects {\small $I^{\text{and}}_{T, v_1}$} and {\small $I^{\text{and}}_{T, v_2}$} can also be merged as {\small $\forall T\subseteq N, I^{\text{and}}_{T, v}=I^{\text{and}}_{T, v_1} + I^{\text{and}}_{T, v_2}$}.

(3) \textit{Dummy axiom}. If a variable {\small $i\in N$} is a dummy variable, \emph{i.e.} {\small $\forall S\subseteq N\setminus\{i\}, v(\mathbf{x}_{S\cup\{i\}})=v(\mathbf{x}_S)+v(\mathbf{x}_{\{i\}})$}, then it has no interaction with other variables, {\small $\forall \ \emptyset\not= T\subseteq N\setminus\{i\}$, $I^{\text{and}}_{T\cup\{i\}}=0$}.

(4) \textit{Symmetry axiom}. If input variables {\small $i,j\in N$} cooperate with other variables in the same way, {\small $\forall S\subseteq N\setminus\{i,j\}, v(\mathbf{x}_{S\cup\{i\}})=v(\mathbf{x}_{S\cup\{j\}})$}, then they have same interaction effects with other variables, {\small $\forall T\subseteq N\setminus\{i,j\}, I^{\text{and}}_{T\cup\{i\}} = I^{\text{and}}_{T\cup\{j\}}$}.

(5) \textit{Anonymity axiom}. For any permutations $\pi$ on {\small $N$}, we have {\small $\forall T \!\subseteq\! N, I^{\text{and}}_{T, v}=I^{\text{and}}_{\pi T, \pi v}$}, where {\small $\pi T \overset{\text{def}}{=} \{\pi(i) | i \in T\}$}, and the new model {\small $\pi v$} is defined by {\small $(\pi v)(\mathbf{x}_{\pi S}) = v(\mathbf{x}_S)$}. This indicates that interaction effects are not changed by permutation.

(6) \textit{Recursive axiom}. The interaction effects can be computed recursively. For {\small $i\in N$} and {\small $T\subseteq N\setminus\{i\}$}, the interaction effect of the pattern {\small $T\cup\{i\}$} is equal to the interaction effect of {\small $T$} with the presence of $i$ minus the interaction effect of $T$ with the absence of $i$, \emph{i.e.} {\small $\forall T\!\subseteq\! N\!\setminus\!\{i\}, I^{\text{and}}_{T\cup \{i\}}=I^{\text{and}}_{T, i\text{ present}} - I^{\text{and}}_T$}. {\small $I^{\text{and}}_{T, i\text{ present}}$} denotes the interaction effect when the variable $i$ is always present as a constant context, \emph{i.e.} {\small $I^{\text{and}}_{T, i\text{ present}}=\sum_{L\subseteq T} (-1)^{|T|-|L|}\cdot v(\mathbf{x}_{L\cup\{i\}})$}.

(7) \textit{Interaction distribution axiom}. This axiom characterizes how interactions are distributed for ``interaction functions''~\cite{sundararajan2020shapley}. An interaction function {\small $v_T$} parameterized by a subset of variables {\small $T$} is defined as follows. {\small $\forall S\subseteq N$}, if {\small $T\subseteq S$}, {\small$v_T(\mathbf{x}_S)=c$} ; otherwise, {\small $v_T(\mathbf{x}_S)=0$}. The function {\small$v_T$} models pure interaction among the variables in {\small$T$}, because only if all variables in {\small$T$} are present, the output value will be increased by {\small$c$}. The interactions encoded in the function {\small$v_T$} satisfies {\small $I^{\text{and}}_T=c$}, and {\small $\forall S\neq T$}, {\small $I^{\text{and}}_S=0$}.


\section{Common conditions for sparse interactions}
\label{sec:apdx-condition-for-sparsity}

\citet{ren2024we} have proved three sufficient conditions for the sparsity of AND interactions.

\textbf{Condition 1.} \textit{The DNN does not encode extremely high-order interactions: {\small$\forall \ T\in \{T\subseteq N \mid \vert T\vert \ge M+1\}, \ I^{\text{\rm and}}_T =0$}.}

Condition 1 is common because extremely high-order interactions usually represent very complex and over-fitted patterns, which are unlikely to be learned by a well-trained DNN in real scenarios.

\textbf{Condition 2.} \textit{Let {\small$\bar{u}^{(k)}\overset{\text{\rm def}}{=}\mathbb{E}_{|S|=k}[v(\mathbf{x}_S)-v(\mathbf{x}_\emptyset)]$} denote the average classification confidence of the DNN over all masked samples $\mathbf{x}_S$ with $k$ unmasked input variables. This average classification confidence monotonically increases when $k$ increases: $\forall \ k' \le k$, {\small$\bar{u}^{(k')} \le \bar{u}^{(k)}$}.}

Condition 2 implies that a well-trained DNN is likely to have higher average classification confidence for less masked input samples.

\textbf{Condition 3.} \textit{Given the average classification confidence $\bar{u}^{(k)}$ of samples with $k$ unmasked input variables, there is a polynomial lower bound for the average classification confidence with $k' (k'\le k)$ unmasked input variables: {\small $\forall \ k' \le k, \ \bar{u}^{(k')} \ge (\frac{k'}{k})^p \ \bar{u}^{(k)}$}, where $p>0$ is a constant.}

Condition 3 suggests that the classification confidence of the DNN remains relatively stable even when presented with masked input samples. In real-world applications, the classification or detection of masked or occluded samples frequently occurs. As a result, a well-trained DNN typically develops the ability to classify such masked inputs by leveraging local information, which can be derived from the visible portions of the input. Consequently, the model should not produce a substantially reduced confidence score for masked samples.


\section{Details to extract the sparsest AND-OR interactions}
\label{sec:apdx-optimize-pq}

A method is proposed~\cite{li2023defining, chen2024defining} to simultaneously extract AND interactions $I^{\text{and}}_T$ and OR interactions $I^{\text{or}}_T$ from the network output. Given a masked sample $\mathbf{x}_L$, \cite{li2023defining} proposed to learn a decomposition $v(\mathbf{x}_L)=u^{\text{and}}_L + u^{\text{or}}_L$ towards the sparsest interactions.
The component {$u^{\text{and}}_L$} was explained by AND interactions, and the component {$u^{\text{or}}_L$} was explained by OR interactions.
Specifically, they decomposed $v(\mathbf{x}_L)$ into $u^{\text{and}}_L= 0.5 \cdot v(\mathbf{x}_L)+\gamma_L$ and $u^{\text{or}}_L= 0.5 \cdot v(\mathbf{x}_L) -\gamma_L$, where $\{\gamma_L:L\subseteq N\}$ is a set of learnable variables that determine the decomposition. In this way, the AND interactions and OR interactions can be computed according to Theorem~\ref{theorem:match}, \textit{i.e.}, $I^{\text{and}}_T=\sum\nolimits_{L \subseteq T}(-1)^{|T|-|L|} u^{\text{and}}_L$, and  $I^{\text{or}}_T=-\sum\nolimits_{L \subseteq T}(-1)^{|T|-|L|} v^{\text{or}}_{N \setminus L}$.

The parameters $\{\gamma_L\}$ were learned by minimizing the following LASSO-like loss to obtain sparse interactions:
\begin{equation}
\label{eq:loss-pq}
    \min_{\{\gamma_L\}} \sum_{T\subseteq N} \vert I^{\text{and}}_T \vert + \vert I^{\text{or}}_T \vert
\end{equation}

\textbf{Removing small noises.} A small noise $\delta$ in the network output may significantly affect the extracted interactions, especially for high-order interactions. Thus, ~\cite{li2023defining} proposed to learn to remove a small noise term $\delta_T$ from the computation of AND-OR interactions.
Specifically, the decomposition was rewritten as {$u^{\text{and}}_L=0.5 (v(\mathbf{x}_L) -\delta_L) +\gamma_L$} and {$u^{\text{or}}_L=0.5 (v(\mathbf{x}_L)-\delta_L) +\gamma_L$}.
Thus, the parameters {$\{\delta_L\}$} and {$\{\gamma_L\}$} are simultaneously learned by minimizing the loss function in Eq.~(\ref{eq:loss-pq}).
The values of {$\{\delta_L\}$} were constrained in $[-\zeta, \zeta]$ where {$\zeta=0.02\cdot \vert v(\mathbf{x})-v(\mathbf{x}_\emptyset) \vert$}.


\section{Proof of Theorem \ref{theorem:match}}
\label{proof:match}

\begin{proof} \textbf{(1) Universal matching theorem of AND interactions.}

We will prove that output component \( v^{\text{and}}_S \) on all \( 2^n \) masked samples \( \{\mathbf{x}_S:S\subseteq N\} \) could be universally explained by the all interactions in \( S\subseteq N \), \emph{i.e.}, \( \forall \emptyset \neq S\subseteq N, v^{\text{and}}_S = \sum_{\emptyset \neq T\subseteq S} I^{\text{and}}_T + v(\mathbf{x}_\emptyset) \). In particular, we define \( v^{\text{and}}_\emptyset = v(\mathbf{x}_\emptyset) \) (\textit{i.e.}, we attribute output on an empty sample to AND interactions).

Specifically, the AND interaction is defined as \( I^{\text{and}}_T = \sum\nolimits_{L \subseteq T} (-1)^{|T|-|L|} u^{\text{and}}_L \).
To compute the sum of AND interactions \( \sum_{\emptyset \neq T\subseteq S} I^{\text{and}}_T = \sum\nolimits_{\emptyset \neq T \subseteq S} \sum\nolimits_{L \subseteq T} (-1)^{\vert T \vert - \vert L \vert} u^{\text{and}}_L \), we first exchange the order of summation of the set \( L\subseteq T\subseteq S \) and the set \( T \supseteq L \).
That is, we compute all linear combinations of all sets \( T \) containing \( L \) with respect to the model outputs \( u^{\text{and}}_L \) given a set of input variables \( L \), \textit{i.e.}, \( \sum\nolimits_{T: L \subseteq T \subseteq S} (-1)^{|T|-|L|} u^{\text{and}}_L \).
Then, we compute all summations over the set \( L\subseteq S \).

In this way, we can compute them separately for different cases of \( L\subseteq T\subseteq S \). In the following, we consider the cases (1) \( L = S = T \), and (2) \( L\subseteq T\subseteq S, L\ne S \), respectively.

(1) When \( L=S=T \), the linear combination of all subsets \( T \) containing \( L \) with respect to the model output \( u^{\text{and}}_L \) is \( (-1)^{|S|-|S|} u^{\text{and}}_L = u^{\text{and}}_L \).

(2) When \( L\subseteq T\subseteq S, L\ne S \), the linear combination of all subsets \( T \) containing \( L \) with respect to the model output \( u^{\text{and}}_L \) is \( \sum\nolimits_{T: L \subseteq T \subseteq S} (-1)^{|T|-|L|} u^{\text{and}}_L \). For all sets \( T: S\supseteq T\supseteq L \), let us consider the linear combinations of all sets \( T \) with number \( |T| \) for the model output \( u^{\text{and}}_L \), respectively. Let \( m := |T| - |L| \), (\( 0\le m\le |S|-|L| \)), then there are a total of \( C_{|S|-|L|}^{m} \) combinations of all sets \( T \) of order \( |T| \). Thus, given \( L \), accumulating the model outputs \( u^{\text{and}}_L \) corresponding to all \( T\supseteq L \), then \( \sum\nolimits_{T: L \subseteq T \subseteq S} (-1)^{|T|-|L|} u^{\text{and}}_L = u^{\text{and}}_L \cdot \underbrace{\sum\nolimits_{m=0}^{\vert S \vert - \vert L \vert} C_{|S|-|L|}^m (-1)^m}_{=0} = 0 \). Please see the complete derivation of the following formula.

\begin{equation}\begin{aligned}
    \sum\nolimits_{\emptyset \neq T \subseteq S} I^{\text{and}}_T
    = &  \sum\nolimits_{\emptyset \neq T \subseteq S} \sum\nolimits_{L \subseteq T} (-1)^{\vert T \vert - \vert L \vert} u^{\text{and}}_L \\
    = & \sum\nolimits_{L \subseteq S} \sum\nolimits_{T: L \subseteq T \subseteq S} (-1)^{\vert T \vert - \vert L \vert} u^{\text{and}}_L  - v^{\text{and}}_\emptyset \\
    = & \underbrace{v^{\text{and}}_S}_{L = S} + \sum\nolimits_{L \subseteq S, L \neq S} u^{\text{and}}_L \cdot \underbrace{\sum\nolimits_{m=0}^{\vert S \vert - \vert L \vert} C_{|S|-|L|}^m (-1)^m}_{=0}   - v^{\text{and}}_\emptyset \\
     = & v^{\text{and}}_S  - v^{\text{and}}_\emptyset  = v^{\text{and}}_S  - v(\mathbf{x}_\emptyset)
\end{aligned}\end{equation}

Thus, we have \( \forall \emptyset \neq S\subseteq N, v^{\text{and}}_S = \sum_{\emptyset \neq T\subseteq S} I^{\text{and}}_T + v(\mathbf{x}_\emptyset) \).

\textbf{(2) Universal matching theorem of OR interactions.}

According to the definition of OR interactions, we will derive that \( \forall S\subseteq N, v^{\text{or}}_S = \sum_{T:T\cap S\neq \emptyset} I^{\text{or}}_T \),
where we define \( v^{\text{or}}_\emptyset = 0 \) (recall that in Step (1), we attribute the output on empty input to AND interactions).

Specifically, the OR interaction is defined as \( I^{\text{or}}_T = -\sum\nolimits_{L \subseteq T} (-1)^{|T|-|L|} v^{\text{or}}_{N\setminus L} \).
Similar to the above derivation of the universal matching theorem of AND interactions, to compute the sum of OR interactions \( \sum\nolimits_{T:T \cap S \neq \emptyset} I^{\text{or}}_T = \sum\nolimits_{T:T \cap S \neq \emptyset} \left[- \sum\nolimits_{L \subseteq T} (-1)^{\vert T \vert - \vert L \vert} v^{\text{or}}_{N \setminus L} \right] \), we first exchange the order of summation of the set \( L\subseteq T \subseteq N \) and the set \( T:T \cap S \neq \emptyset \). That is, we compute all linear combinations of all sets \( T \) containing \( L \) with respect to the model outputs \( v^{\text{or}}_{N \setminus L} \) given a set of input variables \( L \), \textit{i.e.}, \( \sum\nolimits_{T: T \cap S \neq \emptyset, T \supseteq L} (-1)^{\vert T \vert - \vert L \vert} v^{\text{or}}_{N \setminus L} \). Then, we compute all summations over the set \( L\subseteq N \).

In this way, we can compute them separately for different cases of \( L\subseteq T\subseteq N, T \cap S \neq \emptyset \). In the following, we consider the cases (1) \( L = N \setminus S \), (2) \( L=N \), (3) \( L \cap S \neq \emptyset, L \neq N \), and (4) \( L \cap S=\emptyset, L \neq N \setminus S \), respectively.

(1) When \( L = N \setminus S \), the linear combination of all subsets \( T \) containing \( L \) with respect to the model output \( v^{\text{or}}_{N \setminus L} \) is \( \sum\nolimits_{T: T \cap S \neq \emptyset, T \supseteq L} (-1)^{\vert T \vert - \vert L \vert} v^{\text{or}}_{N \setminus L} = \sum\nolimits_{T: T \cap S \neq \emptyset, T \supseteq L} (-1)^{\vert T \vert - \vert L \vert} v^{\text{or}}_S \). For all sets \( T: T\supseteq L, T \cap S \neq \emptyset \) (then \( T \neq N \setminus S, T \neq L \)), let us consider the linear combinations of all sets \( T \) with number \( |T| \) for the model output \( v^{\text{or}}_S \), respectively. Let \( |T'| := |T| - |L| \), (\( 1\le |T'|\le |S| \)), then there are a total of \( C_{|S|}^{|T'|} \) combinations of all sets \( T' \) of order \( |T'| \).
Thus, given \( L \), accumulating the model outputs \( v^{\text{or}}_S \) corresponding to all \( T\supseteq L \), then \( \sum\nolimits_{T: T \cap S \neq \emptyset, T \supseteq L} (-1)^{\vert T \vert - \vert L \vert} v^{\text{or}}_{N \setminus L} = v^{\text{or}}_S \cdot \underbrace{\sum\nolimits_{|T'|=1}^{\vert S \vert } C_{|S|}^{|T'|} (-1)^{|T'|}}_{=-1} = -v^{\text{or}}_S \).

(2) When \( L=N \) (then \( T=N \)), the linear combination of all subsets \( T \) containing \( L \) with respect to the model output \( v^{\text{or}}_{N \setminus L} \) is \( \sum\nolimits_{T: T \cap S \neq \emptyset, T \supseteq L} (-1)^{\vert T \vert - \vert L \vert} v^{\text{or}}_{N \setminus L} = (-1)^{\vert N \vert - \vert N \vert} v^{\text{or}}_\emptyset = v^{\text{or}}_\emptyset \).

(3) When \( L \cap S \neq \emptyset, L \neq N \), the linear combination of all subsets \( T \) containing \( L \) with respect to the model output \( v^{\text{or}}_{N \setminus L} \) is \( \sum\nolimits_{T: T \cap S \neq \emptyset, T \supseteq L} (-1)^{\vert T \vert - \vert L \vert} v^{\text{or}}_{N \setminus L} \). For all sets \( T: T\supseteq L, T \cap S \neq \emptyset \), let us consider the linear combinations of all sets \( T \) with number \( |T| \) for the model output \( v^{\text{or}}_S \), respectively. Let us split \( |T| - |L| \) into \( |T'| \) and \( |T''| \), \textit{i.e.}, \( |T| - |L| = |T'| + |T''| \), where \( T'=\{i|i\in T, i\notin L, i\in N\setminus S\} \), \( T''=\{i|i\in T, i\notin L, i\in S\} \) (then \( 0\le|T''|\le|S|-|S\cap L| \)) and \( |T'| + |T''| + |L| = |T| \). In this way, there are a total of \( C_{|S|-|S\cap L|}^{|T''|} \) combinations of all sets \( T'' \) of order \( |T''| \). Thus, given \( L \), accumulating the model outputs \( v^{\text{or}}_{N\setminus L} \) corresponding to all \( T\supseteq L \), then \( \sum\nolimits_{T: T \cap S \neq \emptyset, T \supseteq L} (-1)^{\vert T \vert - \vert L \vert} v^{\text{or}}_{N \setminus L} = v^{\text{or}}_{N \setminus L} \cdot \sum_{T' \subseteq N\setminus S \setminus L} \underbrace{\sum\nolimits_{\vert T'' \vert = 0}^{\vert S \vert-\vert S \cap L \vert} C_{\vert S \vert - \vert S \cap L \vert}^{\vert T''\vert } (-1)^{\vert T' \vert + \vert T'' \vert} }_{=0} = 0 \).

(4) When \( L \cap S=\emptyset, L \neq N \setminus S \), the linear combination of all subsets \( T \) containing \( L \) with respect to the model output \( v^{\text{or}}_{N \setminus L} \) is \( \sum\nolimits_{T: T \cap S \neq \emptyset, T \supseteq L} (-1)^{\vert T \vert - \vert L \vert} v^{\text{or}}_{N \setminus L} \). Similarly, let us split \( |T| - |L| \) into \( |T'| \) and \( |T''| \), \textit{i.e.}, \( |T| - |L| = |T'| + |T''| \), where \( T'=\{i|i\in T, i\notin L, i\in N\setminus S\} \), \( T''=\{i|i\in T, i\in S\} \) (then \( 0\le|T''|\le|S| \)) and \( |T'| + |T''| + |L| = |T| \). In this way, there are a total of \( C_{|S|}^{|T''|} \) combinations of all sets \( T'' \) of order \( |T''| \). Thus, given \( L \), accumulating the model outputs \( v^{\text{or}}_{N\setminus L} \) corresponding to all \( T\supseteq L \), then \( \sum\nolimits_{T: T \cap S \neq \emptyset, T \supseteq L} (-1)^{\vert T \vert - \vert L \vert} v^{\text{or}}_{N \setminus L} = v^{\text{or}}_{N \setminus L} \cdot \sum_{T' \subseteq N\setminus S \setminus L} \underbrace{\sum\nolimits_{\vert T'' \vert = 0}^{\vert S \vert} C_{\vert S \vert }^{\vert T''\vert } (-1)^{\vert T' \vert + \vert T'' \vert} }_{=0} = 0 \).

Please see the complete derivation of the following formula.
\begin{equation}\begin{small}
\begin{aligned}
\sum\nolimits_{T:T \cap S \neq \emptyset} I^{\text{or}}_T
        &= \sum\nolimits_{T:T \cap S \neq \emptyset} \left[- \sum\nolimits_{L \subseteq T} (-1)^{\vert T \vert - \vert L \vert} v^{\text{or}}_{N \setminus L} \right]\\
        &= - \sum\nolimits_{L \subseteq N} \sum\nolimits_{T: T \cap S \neq \emptyset, T \supseteq L} (-1)^{\vert T \vert - \vert L \vert} v^{\text{or}}_{N \setminus L} \\
        &=  - \left[\sum_{\vert T' \vert = 1}^{\vert S \vert} C_{\vert S \vert}^{\vert T' \vert} (-1)^{\vert T' \vert} \right] \cdot \underbrace{v^{\text{or}}_S}_{L=N\setminus S} - \underbrace{v^{\text{or}}_\emptyset}_{L=N} \\
        &\quad- \sum_{L \cap S \neq \emptyset, L \neq N} \left[\sum_{T' \subseteq N\setminus S \setminus L} \left( \sum_{\vert T'' \vert = 0}^{\vert S \vert-\vert S \cap L \vert} C_{\vert S \vert - \vert S \cap L \vert}^{\vert T''\vert } (-1)^{\vert T' \vert + \vert T'' \vert} \right) \right]\cdot v^{\text{or}}_{N \setminus L}  \\
        &\quad- \sum_{L \cap S=\emptyset, L \neq N \setminus S} \left[ \sum_{T' \subseteq N\setminus S \setminus L} \left( \sum_{\vert T'' \vert=0}^{\vert S \vert} C_{\vert S \vert}^{\vert T'' \vert} (-1)^{\vert T' \vert + \vert T'' \vert}\right) \right] \cdot v^{\text{or}}_{N \setminus L}  \\
        &=  - (-1) \cdot v^{\text{or}}_S - v^{\text{or}}_\emptyset - \sum_{L \cap S \neq \emptyset, L \neq N} \left[\sum_{T' \subseteq N\setminus S \setminus L} 0 \right]\cdot v^{\text{or}}_{N \setminus L}  \\
        &\quad- \sum_{L \cap S=\emptyset, L \neq N \setminus S}\left[\sum_{T' \subseteq N\setminus S \setminus L} 0 \right] \cdot v^{\text{or}}_{N \setminus L}  \\
        &= v^{\text{or}}_S - v^{\text{or}}_\emptyset\\
        &= v^{\text{or}}_S
\end{aligned} \end{small}
\end{equation}

\textbf{(3) Universal matching theorem of AND-OR interactions.}

With the universal matching theorem of AND interactions and the universal matching theorem of OR interactions, we can easily get \( v(\mathbf{x}_S) = v^{\text{and}}_S + v^{\text{or}}_S
= v(\mathbf{x}_\emptyset) + \sum_{\emptyset \neq T\subseteq S} I^{\text{and}}_T + \sum_{T: T\cap S \neq \emptyset} I^{\text{or}}_T \), thus, we obtain the universal matching theorem of AND-OR interactions.

\end{proof}


\section{Experimental detail}\label{sec:experimental_setting}
\subsection{Training settings}
In this paper, we trained various DNNs for different tasks. Specifically, for the image classification task, we trained VGG-11/13~\cite{simonyan2014very} on the MNIST dataset~\cite{lecun1998gradient} with a learning rate of $0.01$. We trained VGG-11/13 on the CIFAR-10 dataset~\cite{krizhevsky2009learning} with a learning rate of $0.01$.  We trained ResNet-20 on the CIFAR-10 dataset and MNIST dataset. We trained VGG-16 on the CUB200-2011 dataset~\cite{wah2011caltech} (using bird images cropped from the background) with a learning rate of $0.01$. We trained AlexNet~\cite{krizhevsky2012imagenet} on the Tiny-ImageNet dataset~\cite{tiny-imagenet} with a learning rate of $0.01$. We trained ResNet56/34 on the CIFAR-10 dataset with a learning rate of $0.001$. For the sentiment classification task, we trained the Bert-Tiny model~\cite{devlin2018bert} on the SST-2 dataset~\cite{socher2013recursive} with a learning rate of $0.01$. All DNNs were trained using the SGD optimizer~\cite{robbins1951stochastic} with a momentum of 0.9.

For partial experiments, we adopted \( \ell_\infty \)-norm bounded adversarial training following the approach of ~\cite{madry2017towards}. Specifically, adversarial examples were generated using a single-step Projected Gradient Descent (PGD) attack with a maximum perturbation size of \( \epsilon = 4/255 \), step size of \( \alpha = 4/255 \), and \( n_{\text{step}} = 1 \). Specifically, apart from the experiments in Section \ref{ref::sec3.2} (where we aimed to explore the changes in the distribution of interactions during the normal training process), all other experiments were conducted with adversarial training.

\begin{figure}[t]
    \centering
    \includegraphics[width=0.98\textwidth, height=8cm]{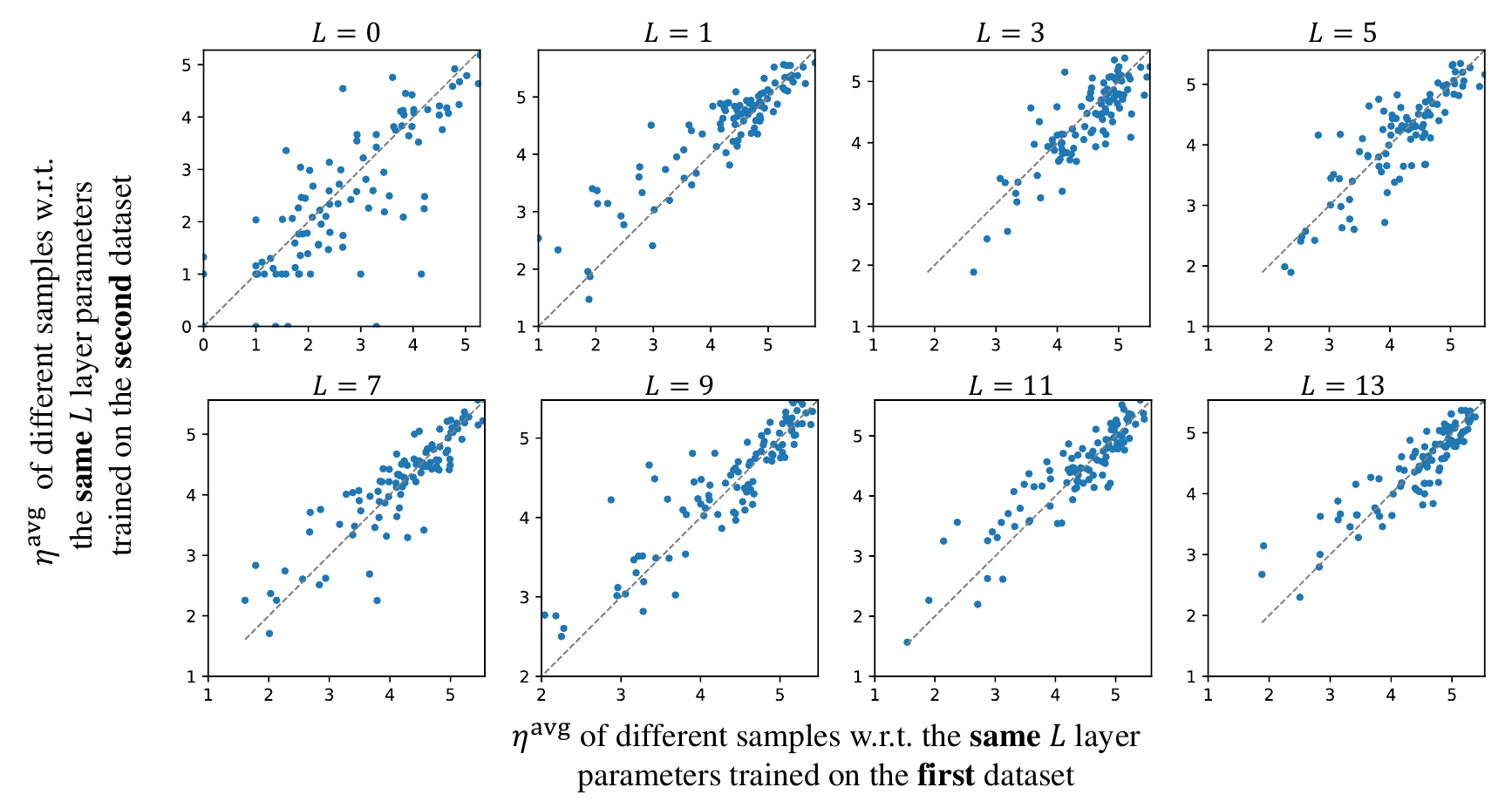}
    \caption{Comparing composition of confusing samples of two DNNs with the \textbf{same} $L$ low-layer parameters, which are trained on \textbf{different} datasets.}
    \label{ref::appendix}
\end{figure}

\subsection{Details about how to calculate interactions for different DNNs}\label{sec:players}
\textbf{\textbullet \ For image data in different image datasets,} since the computational cost of interactions was intolerable, we applied a sampling-based approximation method to calculate AND-OR interactions. Specifically, we considered the feature map after low-layer as intermediate-layer features of DNNs. We uniformly split  the central region of each intermediate-layer feature (\emph{i.e.,} we did not consider the pixel on the edges of an image) into $5 \times 5$ patches and randomly sampled 10 patches to calculate interactions, and considered these patches as input variables for each intermediate-layer feature. We used $\mathbf{0}$ as a baseline value to mask the variables in $N \backslash T$.

\textbf{\textbullet \ For natural language data in SST-2 dataset,} we considered the outputs of the low-layer corresponding to input words as input features. We considered the embeddings corresponding to input features as input variables for each input sentence, and we randomly sampled 10 words, which must have a specific meaning and not be stop words, to calculate interactions. We used the average embedding over different input varibales to mask the tokens in $N \backslash T$.

Specifically, we empirically considered the first 9 convolutional layers of ResNet-56 as the low layers, and considered all the other 47 layers as the high layers. For the Bert-Tiny model, we considered the first transformer block as the low layers, and considered all layers after the first transformer block as the high layers. For other models, we consider the first layer as low layers.

Typically, we compute the mean distribution of interactions over 50 samples.

\subsection{Details on how many epochs the DNN was trained before computing interactions}\label{appendix-sec-epochs}
To compute the Jaccard Similarity of interactions between the training and testing sets, we trained each model for 50 epochs before calculating interactions. Specifically, we trained VGG-13~\citep{simonyan2014very} on the CIFAR-10 dataset~\citep{krizhevsky2009learning}, VGG-11~\citep{simonyan2014very} on the MNIST dataset~\citep{lecun1998gradient}, ResNet-20~\citep{he2016deep} on the CIFAR-10 dataset, and the BERT-Tiny model~\citep{devlin2018bert} on the SST-2 dataset~\citep{socher2013recursive} to evaluate the generalization power of interactions.

To explore the relationship between hard samples and confusing samples, we trained each model for 50 epochs before calculating interactions. Specifically, we trained VGG-11 on the CIFAR-10 dataset, and trained ResNet-20 on the MNIST dataset.

To explore the composition of confusing samples in different DNNs, we trained each model for 200 epochs before calculating interactions. We conducted experiments using ResNet-32 and ResNet-56 trained on the CIFAR-10 dataset.

To explore the impact of a DNN's low-layer parameters, network architecture, and high-layer parameters on the composition of confusing samples, we trained each model for 150 epochs before calculating interactions for ResNet-56 models and trained each model for 10 epochs before calculating interactions for Bert-Tiny models. We trained the ResNet-56 model on the CIFAR-10 dataset and train the Bert-Tiny model on the SST-2 dataset.

\subsection{Details about how to find hard samples for different DNNs}\label{appendix-sec-hard-samples}
To better identify intuitively hard samples, we selected three samples from each class labeled 1 to 9 in the CIFAR-10 and MNIST datasets and reassigned their labels to 0. These samples were more likely to become hard samples, making it easier to compare them with confusing samples and analyze their differences.

\subsection{Details about how to set two DNNs to have the same low-layer parameters}\label{appendix-sec-fix-low-layer-parameters}
In the experiments of Sections \ref{ref::sec4.2}, we explored the impact of different low-layer parameters on the composition of confusing samples in DNNs.  To compare two DNNs trained on same datasets while ensuring they had different low-layer parameters, we replaced the low-layer parameters of the current two DNNs with those from two other well-trained DNNs that had the same architecture but different low-layer parameters.  To compare two DNNs trained on different datasets while ensuring they had identical low-layer parameters, we replaced their low-layer parameters with those from a single well-trained DNN, ensuring consistency in their low-layer parameters.

\begin{figure}[t]
    \centering
    \includegraphics[width=0.78\textwidth, height=10cm]{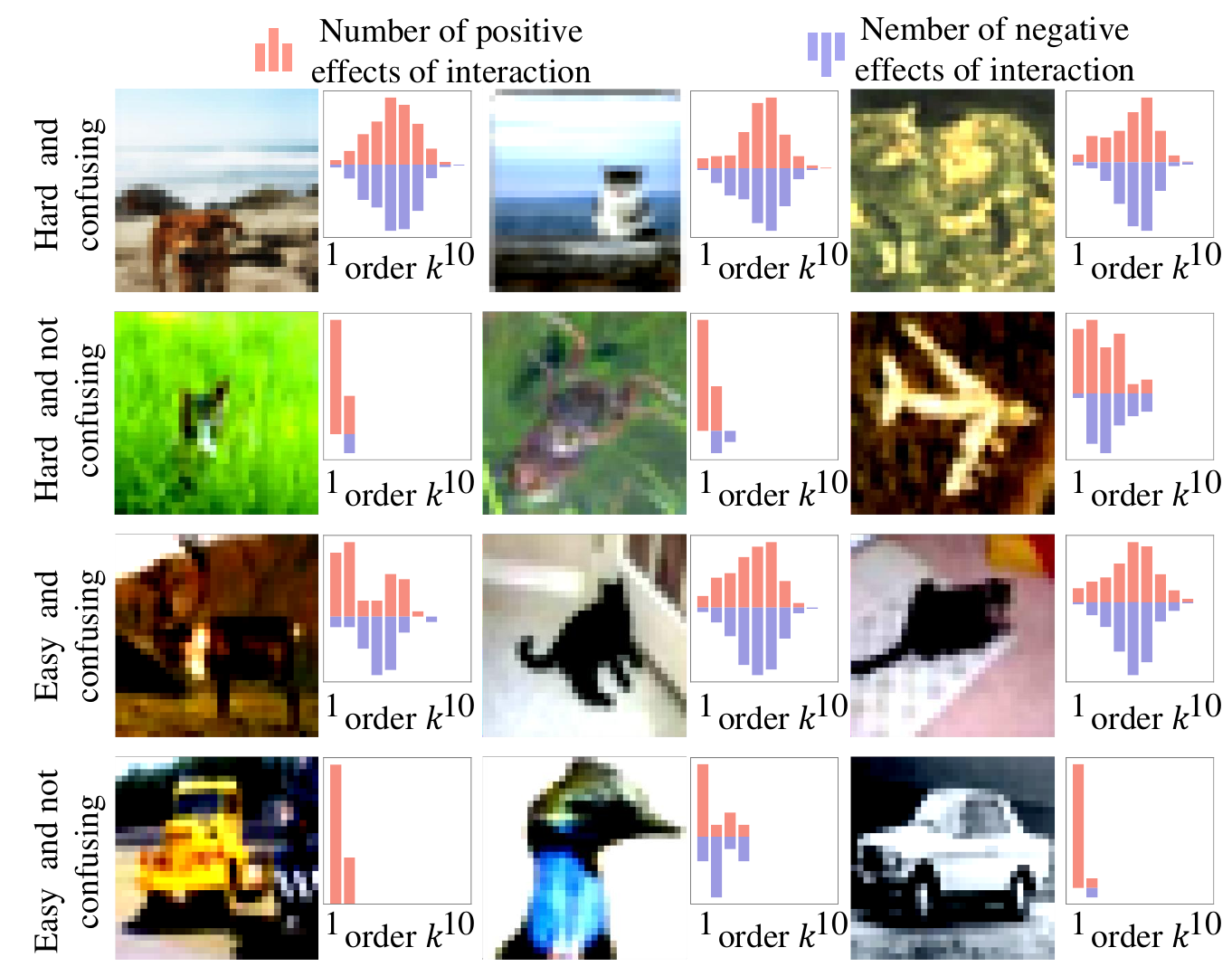}
    \caption{Confusing samples are not same as hard samples. Some hard samples are not confusing samples, and some confusing samples are not hard samples, either.}
    \label{ref::appendix_number}
\end{figure}

\section{More experimental results
}\label{sec:more-result}
\subsection{More results for exploring parameters in how many layers are sufficient to determine confusing samples}
In this subsection, we show more results for exploring parameters in how many layers are sufficient to determine confusing samples. We Compared composition of confusing samples of two DNNs with the same $L$ low-layer parameters, which are trained on different datasets. Specifically, we conducted experiments on $L=0, 1, 3, 5, 7, 9, 11, 13$,  please see Figure \ref{ref::appendix} for details.

\subsection{More results for the number of interactions extracted from hard samples and easy samples}
In this subsection, we show more results for the number of interactions extracted from hard samples and easy samples. Figure \ref{ref::appendix_number} shows that most hard samples encode mutually offsetting interactions, and the other type of hard samples only have a few interactions. \ref{ref::appendix_number} also shows that some confusing but not hard samples contain a large number of interactions, including both lots of mutually offsetting interactions and numerous non-offsetting low-order interactions. In this way, confusing samples are not same as hard samples. Some hard samples are not confusing samples, and some confusing samples are not hard samples, either.


\end{document}